\documentclass[a4paper,fleqn]{cas-sc}

\usepackage[numbers]{natbib}
\def\tsc#1{\csdef{#1}{\textsc{\lowercase{#1}}\xspace}}
\tsc{WGM}
\tsc{QE}
\usepackage{amsmath,amsfonts}
\usepackage{algorithmic}
\usepackage{algorithm}
\usepackage{array}
\usepackage{caption}
\usepackage{multirow}
\usepackage[caption=false,font=normalsize,labelfont=sf,textfont=sf]{subfig}
\usepackage{textcomp}
\usepackage{stfloats}
\usepackage{url}
\usepackage{verbatim}
\usepackage{graphicx}

\begin{document}
\let\WriteBookmarks\relax
\def\floatpagepagefraction{1}
\def\textpagefraction{.001}

\shorttitle{}    

\shortauthors{}  

\title [mode = title]{Instance-aware Image Colorization with Controllable Textual Descriptions and Segmentation Masks}  




\author[1]{Yanru An}[style=chinese, orcid=0009-0004-7470-2045]

\author[1]{Ling Gui}[style=chinese]

\author[2]{Chunlei Cai}[style=chinese]

\author[2]{Tianxiao Ye}[style=chinese]

\author[1]{Jiangchao Yao}[style=chinese]

\author[1]{Guangtao Zhai}[style=chinese, orcid=0000-0001-8165-9322]

\author[1]{Qiang Hu}[style=chinese, orcid=0000-0003-4645-9776
]
\cormark[1]

\author[1]{Xiaoyun Zhang}[style=chinese]
\cormark[1]

% \cormark[1]


% \ead{}



\affiliation[1]{organization={Shanghai Jiao Tong University},
            city={Shanghai},
%          citysep={}, % Uncomment if no comma needed between city and postcode
            postcode={200240}, 
            country={China}}

% Footnote of the second author

% Email id of the second author
% \ead{}



\affiliation[2]{organization={Bilibili Inc},
            city={Shanghai},
%          citysep={}, % Uncomment if no comma needed between city and postcode
            country={China}}

\cortext[1]{Corresponding author}



\begin{abstract}
Recently, the application of deep learning in image colorization has received widespread attention. The maturation of diffusion models has further advanced the development of image colorization models. However, current mainstream image colorization models still face issues such as color bleeding and color binding errors, and cannot colorize images at the instance level. In this paper, we propose a diffusion-based colorization method \textbf{MT-Color} to achieve precise instance-aware colorization with use-provided guidance. To tackle color bleeding issue, we design a pixel-level mask attention mechanism that integrates latent features and conditional gray image features through cross-attention. We use segmentation masks to construct cross-attention masks, preventing pixel information from exchanging between different instances. We also introduce an instance mask and text guidance module that extracts instance masks and text representations of each instance, which are then fused with latent features through self-attention, utilizing instance masks to form self-attention masks to prevent instance texts from guiding the colorization of other areas, thus mitigating color binding errors. Furthermore, we apply a multi-instance sampling strategy, which involves sampling each instance region separately and then fusing the results. Additionally, we have created a specialized dataset for instance-level colorization tasks, \textbf{GPT-color}, by leveraging large visual language models on existing image datasets. Qualitative and quantitative experiments show that our model and dataset outperform previous methods and datasets.
\end{abstract}


% \begin{highlights}
% \item We propose pixel-level masked attention to maintain pixel details in colorization.
% \item We propose instance mask and text guidance to achieve instance-level colorization.
% \item We use multi-instance sampling to precisely colorize instances.
% \item We introduce an automatic pipeline to construct instance-level colorization datasets.
% \end{highlights}

%\nocite{*}

% Keywords

\maketitle

% Main text
\begin{figure}
    \centering
    \includegraphics[width=0.9\textwidth]{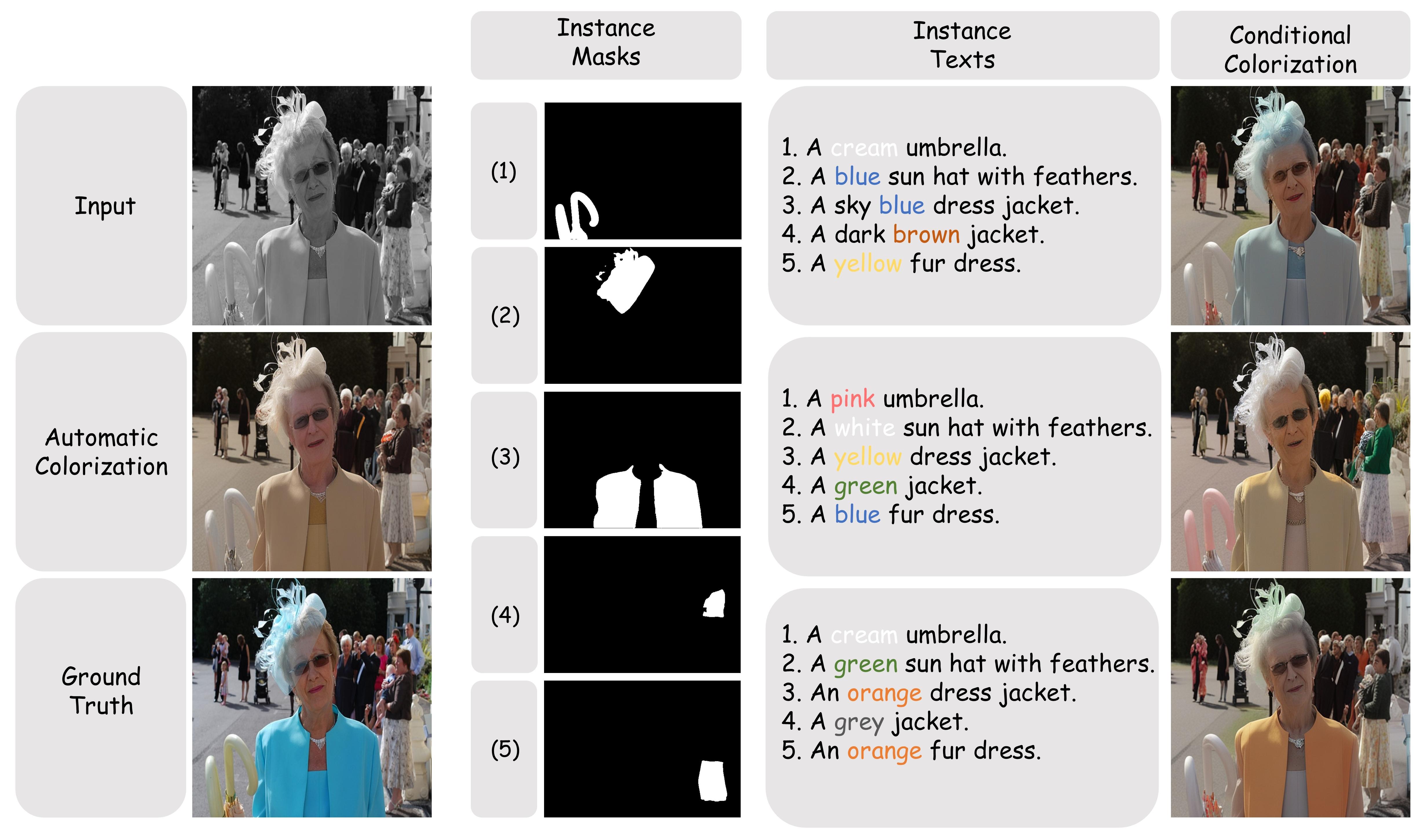} % 插入图片并设置宽度
    \caption{\textbf{MT-Color} can respect: a) generate pleasing unconditional colorization results automatically, b) colorize grayscale images in an instance-aware manner with user-provided instance masks and instance texts. The generation process of MT-Color preserves pixel information and achieve strong color-text binding.}
    \label{teaser}
\end{figure}

\section{Introduction}

Image colorization refers to the process of mapping grayscale images to colorful images. By adding color to grayscale images, image colorization can enhance the information in them and improve visual quality.

In recent years, diffusion probabilistic models \cite{ddpm,ddim} have become one of the most popular research spots. By modeling the reverse process of data structure perturbation through noise and learning from large-scale datasets, diffusion models have achieved powerful and flexible image generation capabilities. Recent works \cite{controlcolor, golocolor} have shown that utilizing pre-trained diffusion model like Stable Diffusion(SD)\cite{sd}'s prior information and ControlNet \cite {controlnet}'s control ability is a viable solution for image colorization.

However, when applied to image colorization tasks, diffusion probabilistic models face the following issues:
\begin{itemize}
    \item \textbf{Color bleeding.} Pretrained Stable Diffusion (SD) models are widely adopted for image colorization tasks due to their strong text-to-image generation priors. However, since the diffusion process in SD is performed in the latent space, it tends to weaken structural and boundary details, often leading to inaccurate pixel reconstruction. Furthermore, the self-attention mechanism in SD computes correlations across all pixel locations, promoting color information exchange between unrelated regions. These limitations frequently result in color bleeding, where the color of one object is influenced by adjacent or unrelated objects.
    \item \textbf{Inaccurate text binding.} The text-guided module of SD uses the CLIP \cite {clip} text encoder to encode text into text embeddings, which are then fused with latent features through cross-attention mechanisms. However, the attention mechanism struggles to effectively identify the correspondence between objects and attributes (e.g., colors) in the text. As a result, when faced with complex textual descriptions, SD may confuse colors between different objects, failing to faithfully restore the text and leading to color binding errors.
    \item \textbf{Sparse color data.} The training datasets for pre-trained diffusion models are often not specifically designed for colorization tasks and lack detailed color information for objects. This results in pre-trained diffusion models being insensitive to the binding relationships between objects and colors in the text, causing mismatches between the colors of objects in the output image and the colors described in the text.
    \item \textbf{Low Resolution.} Diffusion-based image colorization models often struggle to produce high-resolution outputs due to the stochastic nature of the diffusion process and the latent-space denoising used in pre-trained latent diffusion models (LDMs). Although ControlNet introduces additional conditioning from grayscale images, it fails to precisely preserve fine-grained, pixel-level details. As the target resolution increases, colorization results tend to deviate more from the structure of the original grayscale input.

\end{itemize}

In this paper, we propose a novel diffusion-based colorization framework, namely \textbf{MT-Color}. Our method aims to use user-provided instance \textbf{m}asks and instance \textbf{t}exts to achieve precise, instance-aware colorization. MT-Color integrates the powerful generative capability of pre-trained latent diffusion models with the flexible control ability of ControlNet to produce vivid and realistic results.

To mitigate the issue of color bleeding, we propose a \textbf{pixel-level masked attention module} between ControlNet and the U-Net backbone of Stable Diffusion. Specifically, the conditional image features generated by ControlNet are resized and aligned with the U-Net’s latent features via a cross-attention mechanism at the pixel level. To further constrain the attention mechanism, user-provided segmentation masks are employed to restrict the attention regions. This design helps the diffusion model preserve fine-grained spatial details during the generation process. Additionally, by maintaining pixel-level structure, the proposed method enables higher-resolution image generation compared to conventional diffusion-based approaches.

To achieve accurate instance-level colorization and resolve the problem of incorrect color binding, it is crucial to process each instance independently to prevent undesired information exchange. We propose the \textbf{instance mask and text guidance module}, which adds a trainable branch to the self-attention module of U-Net.  This branch jointly encodes instance masks and textual descriptions into instance-specific features, which are then integrated with the latent features via self-attention. The use of instance masks explicitly restricts information flow between different instance regions, alleviating color misbinding. Additionally, we adopt a \textbf{multi-instance sampling strategy}, where the denoising process is performed separately for each instance, further enhancing the instance-awareness of the colorization results.

Additionally, we construct a new dataset, termed GPT-Color, to support the training of our proposed model. We utilize the strong multi-modal reasoning capabilities of pre-trained vision-language model GPT-4\cite{gpt4} and BLIP-2 \cite{blip} to automatically generate high-quality annotations for GPT-color. This dataset provides fine-grained textual descriptions and corresponding segmentation masks for each instance within an image, making it well-suited for the instance-aware colorization task. 

We conduct qualitative and quantitative experiments, along with ablation studies to evaluate the effectiveness of our proposed MT-Color and GPT-color. The results demonstrate that MT-Color produces images that are more perceptually aligned with human expectations compared to existing methods. Moreover, GPT-Color proves to be more effective for the image colorization task than existing datasets.

\section{Related Work}

\subsection{Automatic colorization}
Automatic colorization aims to colorize grayscale images without requiring additional user input. With the advancement of deep learning, data-driven approaches have significantly improved performance.\cite{chengdeepcolorization} first formulate colorization as a regression task using deep networks, while \cite{zhang2016colorfulimagecolorization} cast it as a classification problem.\cite{deshpande2017learningdiverseimagecolorization} adopt a variational autoencoder (VAE) to generate diverse results. To tackle context confusion and edge bleeding, later methods~\cite{zhao2018pixellevelsemanticsguidedimage, zhao2019pixelatedsemanticcolorization} incorporate semantic segmentation. GAN-based approaches such as ChromaGAN~\cite{vitoria2020chromaganadversarialpicturecolorization}, PalGAN~\cite{wang2022palganimagecolorizationpalette}, GCP-Colorization~\cite{wu2022vividdiverseimagecolorization}, and BigColor~\cite{kim2022bigcolorcolorizationusinggenerative} exploit adversarial training to generate vivid images. Recent transformer-based models, including Colorization Transformer~\cite{kumar2021colorizationtransformer}, ColorFormer~\cite{colorformer}, AnchorTransformer~\cite{anchortransformer}, and DDColor~\cite{DDColor}, predict color tokens to produce visually pleasing outputs.

\subsection{Text-based colorization}
Text-based colorization generates plausible colors guided by user-provided textual descriptions. L-CoDeR~\cite{lcoder} introduces a transformer-based framework that unifies image and text modalities and conditions colorization in a coarse-to-fine manner. L-CoIns~\cite{lcoins} enhances instance awareness by incorporating luminance augmentation and a counter-color loss to reduce the correlation between brightness and color words. L-CAD~\cite{lcad} utilizes a pre-trained cross-modal generative model, aligning spatial structures and semantic conditions to achieve instance-aware, text-driven colorization.

\subsection{Diffusion-based colorization}
Diffusion models have shown strong capabilities in image generation~\cite{ddpm, ddim, classifierguidance}. Stable Diffusion~\cite{sd} performs diffusion in latent space, improving efficiency. Works such as GLIDE~\cite{glide} and Imagen~\cite{imagen} leverage pre-trained vision-language models~\cite{clip, t5} for text-guided generation. ControlNet~\cite{controlnet} enables spatial condition control (e.g., edges, depth, segmentation) on pre-trained diffusion models. PASD~\cite{pasd} introduces pixel-aware modules to preserve local structure, benefiting both super-resolution and colorization. Several works\cite{diffusingcolor,piggy,lcad} leverage pre-trained text-to-image diffusion models to achieve text-based colorization. More recently, ControlColor~\cite{controlcolor} addresses color overflow and accuracy issues using self-attention, a deformable autoencoder, and stroke-based color control. GoloColor~\cite{golocolor} extracts global and local embeddings to guide ControlNet-enhanced Stable Diffusion with dense semantic information for precise textual control.

\section{Methodology}
\begin{figure}
    \centering
    \includegraphics[width=1\textwidth]{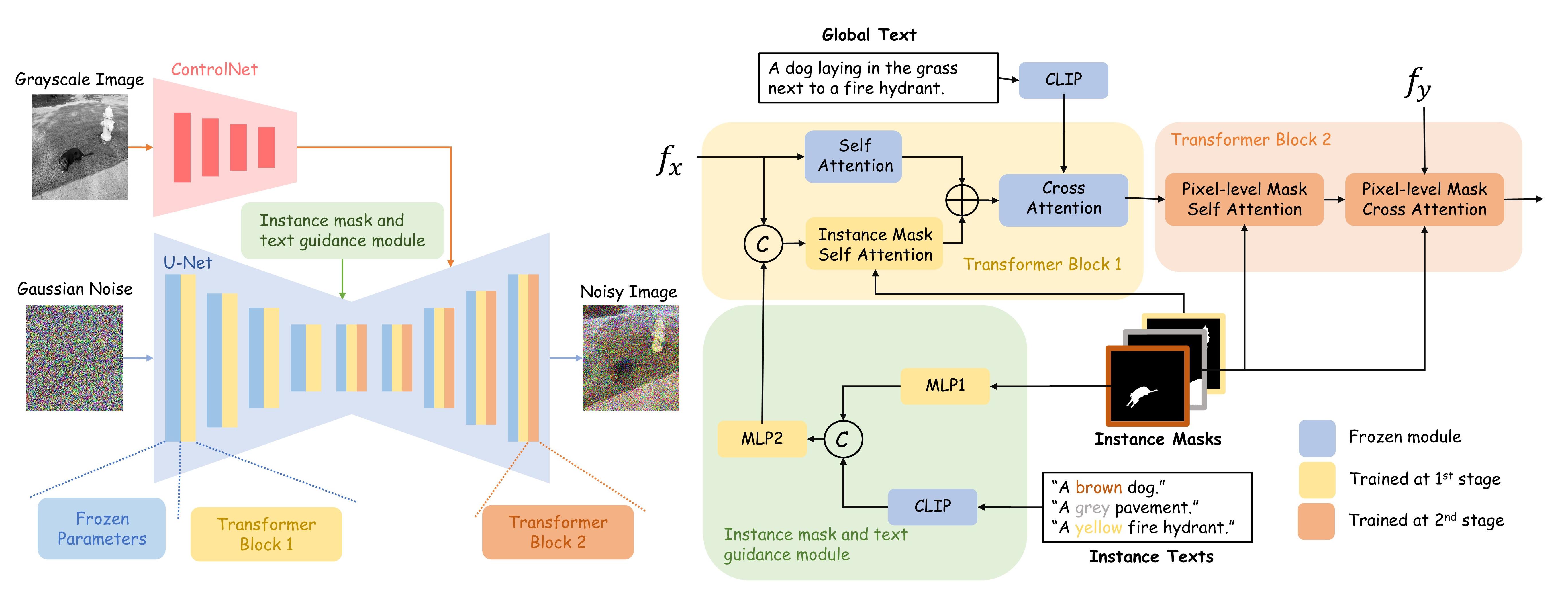} % 插入图片并设置宽度
    \caption{The left shows the overall architecture of our proposed MT-Color, and the right details each module. The instance mask and text guidance module concatenates the feature of instance masks and texts and is connected to the attention module of U-Net. ControlNet is used to extract grayscale image feature, which is integrated with U-Net's latent feature via pixel-level mask attention mechanism.}
    \label{backbone}
\end{figure}

In this section, we first introduce the MT-Color's base method, diffusion models and ControlNet. We then propose the pixel-level masked attention mechanism, which is responsible for the pixel-level fusion of conditional grayscale image representations and latent representations. Next, we detail the instance mask and text guidance module, which integrates instance representations with corresponding tokens in latent features. Moreover, we introduce the multi-instance sampling strategy, which enhances the independence of each instance during the sampling process. Lastly we introduce the vision-language model-aided automatic construction pipeline of our instance-level image colorization dataset, GPT-color.

\subsection{Preliminary}

\subsubsection{Diffusion models}
Diffusion models consist of a forward noising process and a reverse denoising process. In the forward process, Gaussian noise $\epsilon$ is gradually added to the clean data sample $x_0$ over $T$ time steps, resulting in a sequence of progressively noised samples $x_1, \dots, x_T$. The reverse process aims to recover $x_0$ from a noisy input $x_t$ by learning a denoising model $\epsilon_\theta$ that predicts the noise added at each time step $t$. 

To reduce the computational cost of diffusion models in pixel space, the Latent Diffusion Model (LDM) performs the diffusion process in a compressed latent space. Given an optional condition $c$, the training objective of LDM is defined as:
\begin{equation}
    \mathcal{L}_{\text{LDM}} = \mathbb{E}_{x_0, \epsilon, t, c} \left[\left\| \epsilon - \epsilon_\theta(z_t, t, c) \right\|_2^2 \right],
\end{equation}
where $z_t$ denotes the latent representation at time step $t$, and $\epsilon_\theta$ is the denoising network.

\subsubsection{ControlNet}
ControlNet is a neural network architecture designed to introduce explicit conditional control into pretrained text-to-image diffusion models. It constructs a deep and expressive encoder by creating a trainable copy of selected layers from the base LDM. This copy learns to encode additional control signals, while the original model remains mostly fixed. 

The trainable branch and the original model are linked through "zero convolution" layers, which help suppress the propagation of harmful noise during training. The training objective of ControlNet-augmented LDM can be formulated as:
\begin{equation}
    \mathcal{L}_{cond} = \mathbb{E}_{x_0, \epsilon, t, c, y} \left[\left\| \epsilon - \epsilon_\theta(z_t, t, c, y) \right\|_2^2 \right],
\end{equation}
where $c$ denotes the textual condition, and $y$ represents the additional structural condition provided by ControlNet.

\subsection{Pixel-level masked attention mechanism}
As shown in Figure \ref{backbone}, we use pre-trained Latent Diffusion Model(i.e.,SD \cite{sd}) as the backbone, and ControlNet\cite{controlnet} as the conditional grayscale image feature extraction module, which is responsible for integrating the grayscale image feature into the intermediate latent feature of the diffusion backbone. This process transfers the pre-trained diffusion model from the image generation task to the image colorization task.

Although ControlNet supports various types of conditional generation, it cannot utilize grayscale conditional images to achieve precise pixel-level control over the output image, which causes color bleeding issue. To address this issue, we introduce a pixel-level mask attention module between ControlNet and Stable Diffusion's U-Net. One intuitive method is to adjust the size of the conditional image feature output by ControlNet and use a cross-attention mechanism to align it with the latent feature of U-Net at the pixel level, which ensures that the diffusion model faithfully preserves pixel-level details during the diffusion process. However, the direct cross-attention mechanism calculates the correlations between all pixels of the conditional image feature and the latent feature. This implies that pixels of different instances exchange information, which can lead to information leakage between objects, thereby causing color bleeding issues. To address this kind of issue, we introduce instance segmentation masks into the pixel-level attention mechanism, constructing a pixel-level mask attention mechanism.

Specifically, given a latent representation $f_x\in \mathbb{R}^{h\times w\times c}$ of diffusion model and its corresponding conditional image representation $f_y\in \mathbb{R}^{h\times w\times c}$, where $h$, $w$ and $c$ respectively represents the height, width and number of channels of feature maps. And given a set of instance masks $M_n=\{m_k\}_{k=1}^n\in \{0,1\}^{n\times H\times W}$, where $n$, $H$ and $W$ respectively represents the number of instances, height and width of each instance mask. In pixel-level mask attention mechanism, we first adjust the size of $f_x$ and $f_y$ to ${f_x}^\prime \in \mathbb{R}^{h*w\times c}$ and ${f_y}^\prime \in \mathbb{R}^{h*w\times c}$, respectively, and then resize the instance mask set to match the dimensions of the two feature maps, denoted as $M_n^\prime=\{m_k^\prime \}_{k=1}^n\in \{0,1\}^{n\times h\times w}$. For a pixel at position $(i,j), \forall i \in \{1,...w\}, \forall j \in \{1,...,h\}$ in the feature maps, we search within $M_n^\prime$ to find the mask that contains this pixel and then select it as the cross-attention mask for that pixel. After performing this operation to all pixels, we obtain the global cross-attention mask $M$:
\begin{equation}
    M(i,j)=\{m_k^\prime \vert m_k^\prime (i,j)=1,k\in \{1,...,n\}\} \label{mask}
\end{equation}
After copying $M$ to match the number of channels in the feature maps, we compute the output feature map using the cross-attention mechanism and the global mask as follows:
\begin{equation}
    \hat{f_x^\prime}=M\circ\text{Softmax}(\frac{Q^\prime K^{\prime T}}{\sqrt{d}})\cdot V^\prime \label{output feature map}
\end{equation}
where $Q^\prime =W_Q^\prime \cdot f_x^\prime$, $K^\prime =W_K^\prime \cdot f_x^\prime$, $V^\prime =W_V^\prime \cdot f_y^\prime$, and $\circ$ denotes element-wise multiplication of matrices. $W_Q^\prime$ $W_K^\prime$ and $W_V^\prime$ represent learnable projection matrices. Since $f_y^\prime$ is output by ControlNet without undergoing perceptual compression like the autoencoder in Stable Diffusion, it retains the pixel-level details of the conditional grayscale image. By aligning $f_y^\prime$ with the latent feature of U-Net through the pixel-level mask attention mechanism, the diffusion model can acquire the boundary information of the conditional image, and prevents pixels of different objects from exchanging information, thus alleviating the issue of color bleeding.

\subsection{Instance mask and text guidance module}

Many image colorization models do not consider the issue of color binding errors. The few colorization models do address this problem focus their research on cross-attention modules connected to the CLIP \cite{clip} text encoder, and only use instance masks to influence the results. For example, L-CAD \cite{lcad} uses SAM \cite{sam} to segment the mask of every color-described noun in the global text , and uses it as the cross-attention mask for the corresponding color word. GoLoColor\cite{golocolor} fuses the global embedding extracted by BLIP-2 \cite{blip} and the local embedding extracted by RAM\cite{ram} to augment textual control. However, these method does not notice information leakage between instances in self-attention modules. Therefore, we must pay attention to self-attention masks in addition to cross-attention. 

We propose the instance mask and text guidance module, adding a trainable branch to the self-attention module of U-Net. The branch simultaneously uses instance masks and text to influence the results. It encodes them into instance features, which are then imposed with latent features to perform self-attention. Then, by applying instance masks to self-attention layers, it prevent information exchange between pixels in different instance regions, thus addressing the issue of color binding errors. 

Given a latent representation $f_x\in \mathbb{R}^{h\times w\times c}$, a set of instance masks $M_n=\{m_k\}_{k=1}^n\in \{0,1\}^{n\times H\times W}$ and a set of instance texts $T_n=\{\tau_k\}_{k=1}^n\in \mathbb{R}^{n \times l_t}$, where $l_t$ represents the maximum possible length of each instance text, we first convert the instance masks and texts into instance representations that can be input to the instance mask and text guidance module. For instance texts, we use the pre-trained CLIP text encoder to transform $T_n$. For instance masks, we use a multi-layer perceptron (MLP) to extract their features. The MLP consists of 3 convolutional layers. We concatenate the corresponding feature of each instance and pass it through a MLP composed of three fully connected layers. As \ref{instance representation} shows, this process yields the instance feature set $\Gamma_n=\{\gamma_k\}_{k=1}^n\in \mathbb{R}^{n\times l_\gamma}$, where $l_\gamma$is the length of the instance representation.
\begin{equation}
    \Gamma_n=\text{MLP\_2}(\text{concat}(\text{CLIP}(T_n), (\text{MLP\_1}(M_n)))) 
    \label{instance representation}
\end{equation}

Then, we use a masked self-attention mechanism to fuse instance features with latent features from U-Net. We first flatten the latent representation $f_x$, where the flattened feature length is $l_x=h*w$. Next, we concatenate this flattened feature with the instance feature set $\Gamma_n$, yielding a new feature $p \in \mathbb{R}^{l_p \times l_\gamma}$, where the new feature length $l_p=l_x+n$. We then apply a self-attention mechanism to this feature map
\begin{equation}
    \text{self\_map}=\text{Softmax}(\frac{Q_pK_p^T}{\sqrt{d}})\label{selfmap}
\end{equation}
to obtain a self-attention map, denoted as $self\_map$, where $Q_p=W_{Q_p}\cdot p$,$K_p=W_{K_p}\cdot p$,$V_p=W_{V_p}\cdot p$. The size of $self\_map$ is $(l_x+n)\times(l_x+n)$.

We construct self-attention masks with instance mask set. For latent feature's self-attention map \\$self\_map[1:l_x,1:l_x]$, we construct its mask at position $(i,j)$
\begin{equation}
M_{\text{self}}(i,j) =
% \left\{
\begin{cases}
    m_k, &  \exists k \in \{1, \ldots, n\}, m_k(w_i, h_i) = m_k(w_j, h_j) = 1 \\
    0,   & \text{otherwise}
\end{cases}
% \right.
\label{maskself}
\end{equation}
where $(w_i,h_i)$ and $(w_j,h_j)$ are positions in the latent representation corresponding to positions $i$ and $j$ in the self-attention map. If an instance mask includes both $(w_i,h_i)$ and $(w_j, h_j)$, it indicates that the pixels at these two positions belong to the same object, allowing them to exchange information. Conversely, information exchange is prohibited to prevent information leakage.

For the attention map between latent features and instance features $self\_map[1:l_x,l_x+1:l_x+n]$, we similarly construct its mask at the pixel position $(i,j)$,
\begin{equation}
M_{\text{cross}}(i,j) =
\left\{
\begin{aligned}
    & m_j, & \quad m_j(w_i,h_i)=1 \\
    & 0,   & \text{otherwise}
\end{aligned}
\right.
\label{maskcross}
\end{equation}
where, $(w_i,h_i)$ is the position in the latent feature corresponding to position $i$ in the self-attention map.  If the instance mask $m_j$ includes $(w_i,h_i)$, it indicates that the pixel at position $(w_i,h_i)$ in the latent feature is also part of instance $j$. In this case, the pixel at this position can exchange information with the instance; otherwise, it cannot.

We concatenate the self-attention map mask $M_{self}$ of the latent feature with the cross-attention map mask $M_{cross}$ of the instance features along the first dimension, resulting in the complete self-attention map mask $M_{self\_map}$, and use it in the self-attention map to implement the instance mask self-attention mechanism.
\begin{equation}
    \hat{f_x}=(M_\text{self\_map}\circ \text{self\_map}\cdot V_p)[1:l_x,1:l_x]\label{output feature map2}
\end{equation}
Here, we only take the results from the latent feature part, specifically the portion $[1:l_x,1:l_x]$, as the output of the instance mask and text guidance module.

\subsection{Multi-instance sampling for inference}
\begin{figure}
    \centering
    \includegraphics[width=0.6\textwidth]{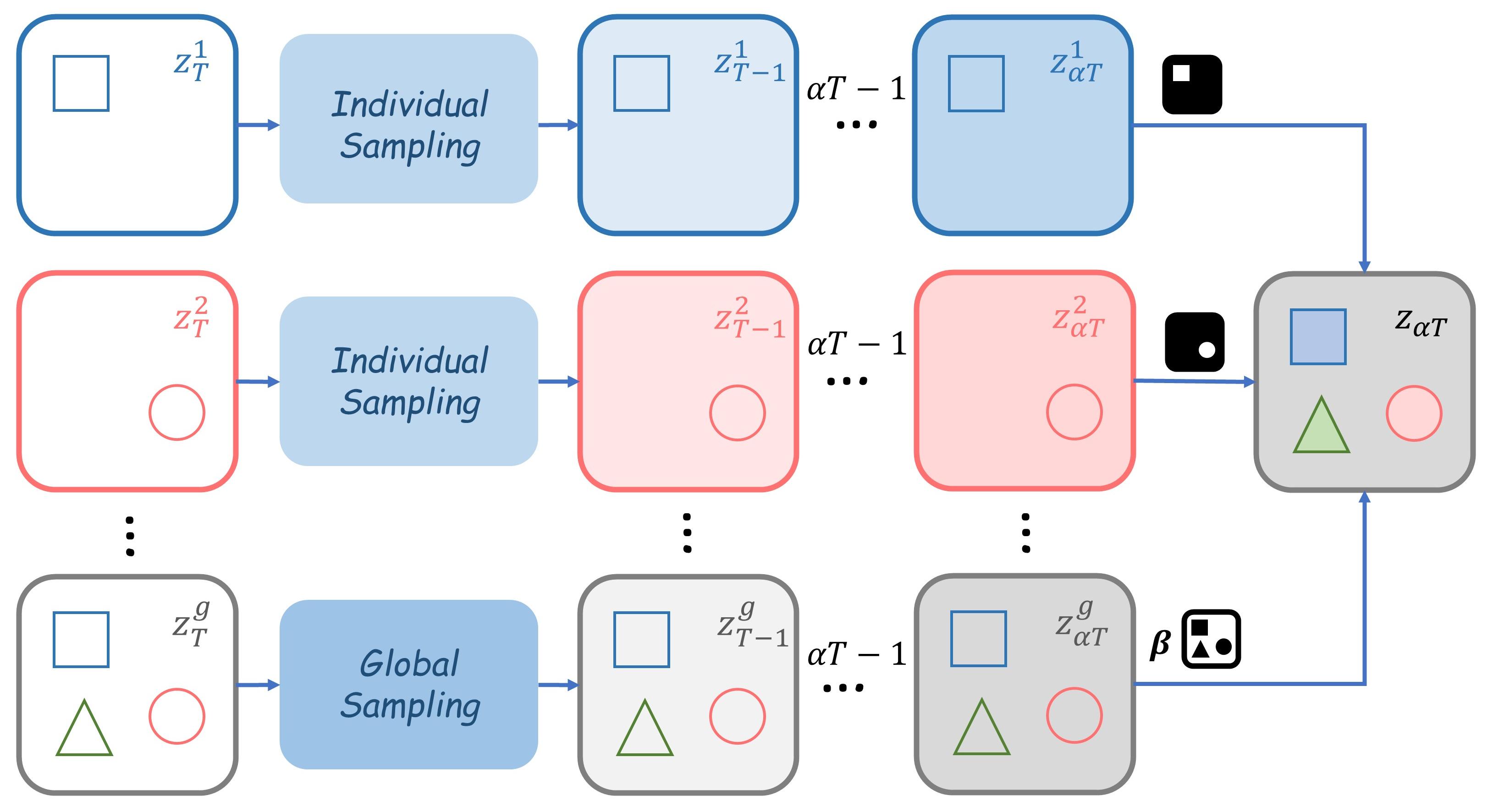} % 插入图片并设置宽度
    \caption{Multi-instance sampling strategy. Instance noises are sampled during the first $\alpha T$ steps, and are cropped and fused together with global noise and then sampled globally in the rest steps.}
    \label{multi-instance sampling}
\end{figure}

Previous work \cite{ella} found that the color information of images generated by diffusion models is determined in the early stage of sampling process. Based on this finding and inspired by the effectiveness of \cite {instancediffusion}, we adopt the multi-instance sampling strategy during model inference to achieve instance-aware colorization. As is illustrated in Figure \ref{multi-instance sampling}, each instance is sampled individually at the beginning of sampling process, taking instance masks and texts as conditions, to obtain instance-specific noisy intermediate images. These images are then weighted and fused with the global noisy intermediate image to serve as input for subsequent sampling steps. 

Specifically, given a series of diffusion steps $\{1,...,T\}$ and an initial Gaussian noise $z_T$ for the global sampling process, we initialize the initial noise for each instance as $z_T^i=z_T,\forall i \in \{1,...,n\}\label{zT}$, where $n$ denotes the number of instances. After colorizing each instance individually, we obtain the set of instance-specific noisy intermediate images $\{z_{\alpha T}^i\}_{i=1}^n$, where $\alpha$ is a hyperparameter that represents the proportion of individual sampling steps to the total sampling steps. Meanwhile, we denoise the global image to obtain the global noisy intermediate image $z_{\alpha T}^g$.

At the end of the individual sampling phase, we perform a weighted fusion of $z_{\alpha T}^g$ and $\{z_{\alpha T}^i\}_{i=1}^n$. We first obtain the global mask through the instance mask set $M_n$: $m_g=\neg \bigvee_{i=1}^n m_i\label{mg}$, where $\bigvee$ denotes the logical OR operation, meaning that all instance masks are combined element-wise into a new mask, and $\neg$ denotes the logical NOT operation, meaning that the resulting mask is inverted element-wise to obtain the global mask. Given a hyperparameter $\beta$, which is the weight of the global noisy intermediate image, we obtain the fused noisy intermediate image.
\begin{equation}
    z_{\alpha T}=\beta m_g\circ z_{\alpha T}^g+\sum_{i=1}^n m_i\circ z_{\alpha T}^i \label{zaT}
\end{equation}
Here, we apply the instance masks to the instance-specific noisy intermediate images to extract information from the corresponding instance regions and paste it onto the weighted global noisy intermediate image, which isolates the information of different instances. After obtaining the fused noisy intermediate image, we proceed to the global sampling phase, $\forall t\in \{\alpha T,...,1\}$,
\begin{equation}
    z_{t-1}=\text{Diffusion}(z_t,t,\tau_g,T_n,M_n)\label{zt-1}
\end{equation}
where $\text{Diffusion}$ represents our model and $\tau_g$ is the global text.

\subsection{Dataset construction pipeline}
\begin{figure}
    \centering
    \includegraphics[width=0.5\textwidth]{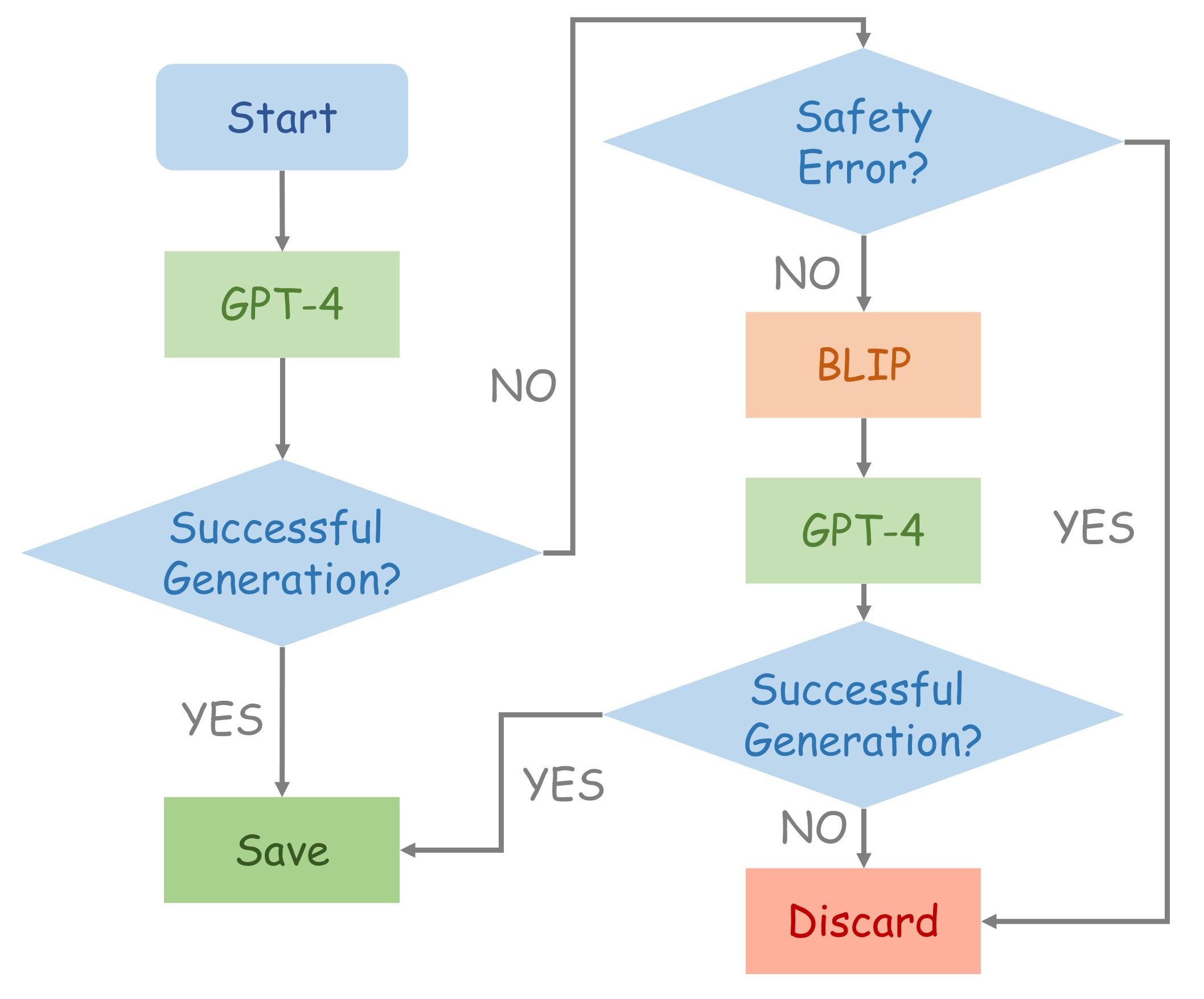} % 插入图片并设置宽度
    \caption{Dataset construction pipeline.}
    \label{gptcolor}
\end{figure}
Currently, mainstream text-based image colorization models are trained using large-scale image datasets like COCO-Stuff\cite{cocostuff} and ImageNet\cite{imagenet}. However, these datasets generally have the following issues:
\begin{itemize}
    \item The image description texts are overly verbose, containing too much information unrelated to image colorization, such as the spatial relationships between objects and the reasons for the scene depicted.
    \item The image description texts do not comprehensively cover the objects and their colors, failing to describe the colors of all objects in the image thoroughly.
    \item There is a lack of individual object descriptions, and the present ones rarely describe the instances’ colors.
\end{itemize}

To address these issues, we want to leverage pre-trained vision-language models designed for image description tasks to generate color-specific texts for both the global image and each instance. For colorization, we only need texts that provide objects and their corresponding colors. Therefore, we believe that an appropriate image colorization dataset should meet the following criteria:
\begin{itemize}
    \item Provide comprehensive descriptive texts that include an global description of the image’s color scheme as well as descriptions of the colors of each object in the image.
    \item Provide segmentation information and texts for each instance in the image, describing each instance in the form of ”object + color” phrases, such as ”a red apple”.
\end{itemize}

First, we utilize the open-source image annotation model RAM\cite{ram} to detect objects in the image and generate their masks and annotations. Next, we selected two leading vision-language models, BLIP-2 and GPT-4, and compared their abilities in generating image texts. When generating instance texts, we use the instance masks to crop out instance images from the global image. We found that the quality of BLIP-2’s descriptions is sometimes inconsistent and includes rare color information, occasionally resulting in failed text generation. We also found that GPT-4 can generally describe objects and their colors well in the format of ”object+color.” However, some images or instances may not pass GPT-4’s safety checks. Additionally, when the input image is blurry or of low resolution, GPT-4 may not generate high-quality descriptions and instead provide invalid text like ”Unable to provide color description, image is too blurred and unclear.” Therefore, we decided to jointly use GPT-4 and BLIP-2 to construct the dataset, as illustrated in Figure \ref{gptcolor}. Based on this pipeline, we construct a dataset specifically for instance-level image colorization tasks, named \textbf{GPT-color}, on a subset of COCO-Stuff. The dataset comprises
approximately 12,000 training images and 3,000 test images. For each image, we provide detailed instance masks and descriptions for an average of 8 instances.

\section{Experiments}
\begin{figure}
    \centering
    \includegraphics[width=1\textwidth]{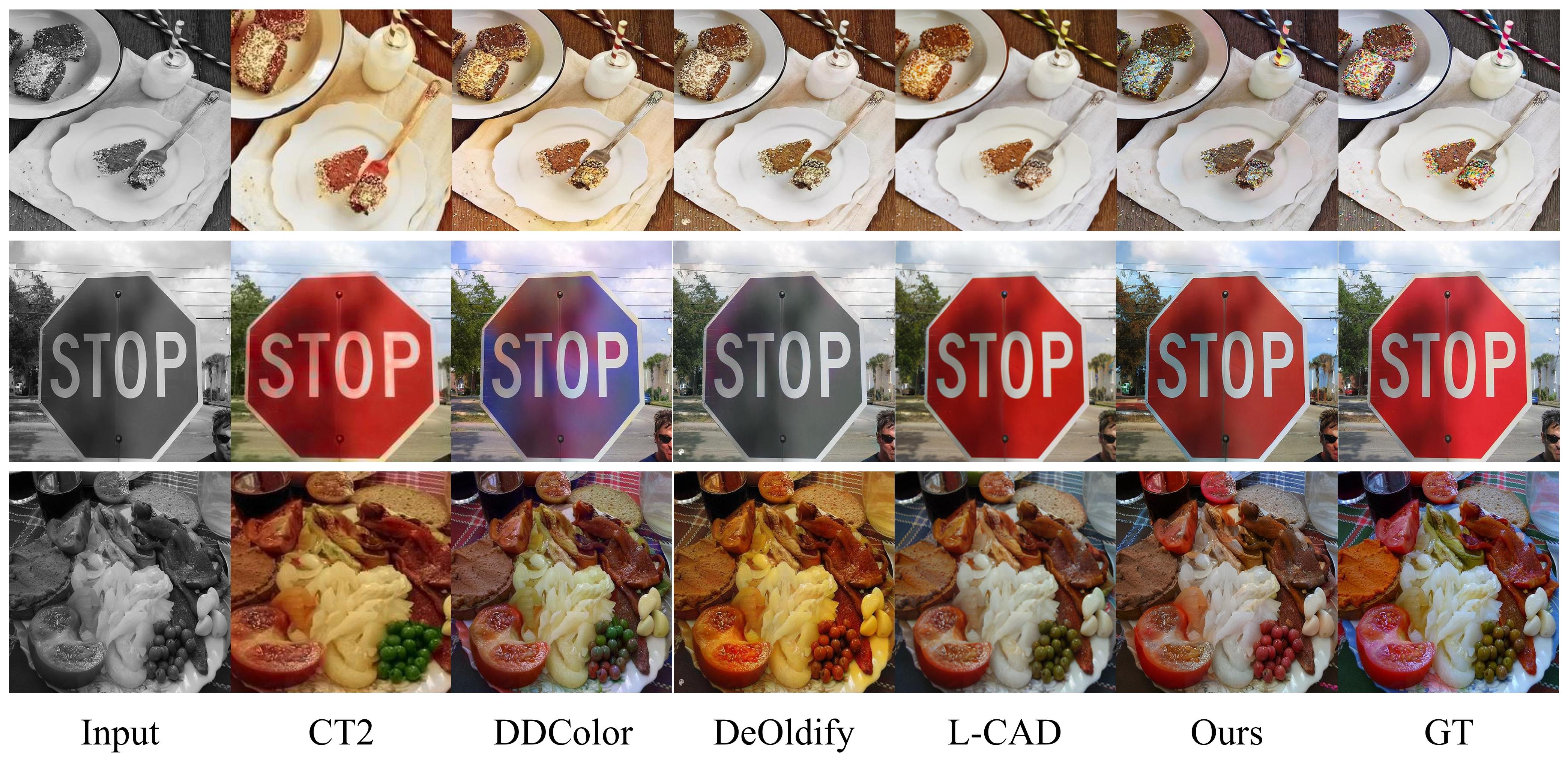} % 插入图片并设置宽度
    \caption{Qualitative comparison results for unconditional colorization. All examples are from GPT-color dataset. Our model generates more human perception-friendly colors and details.}
    \label{qualitative compare}
\end{figure}

\subsection{Training strategy}
Due to the large number of model parameters, direct end-to-end training leads to slow convergence and suboptimal performance. To address this, we adopt a two-stage training strategy.

In the first stage, we train the instance mask and text guidance module independently, as the pixel-level masked attention module in each modified Transformer block relies on its output. In the second stage, we freeze the parameters of the pretrained instance mask and text guidance module, and then introduce ControlNet and the pixel-level masked attention module into the model. Only the parameters of these newly introduced components are updated during this stage.

In both stages, the model is optimized using an L2 loss function defined as:
\begin{equation}
    \mathcal{L}=\mathbb{E}_{x_0, \epsilon, t, c, \tau_g, M_n, T_n}\|\epsilon - \epsilon_\Theta(z_t, t, \tau_g, c, M_n, T_n)\|^2_2
\end{equation}
where $z_t$ denotes the noisy latent representation after $t$ steps of noise addition, $\tau_g$ is the global textual description, $c$ is the conditional grayscale input, and $\Theta$ denotes all trainable parameters.

\subsection{Experiment settings}
We train our model on the GPT-Color dataset using the AdamW\cite{adamw} optimizer. The learning rate is linearly warmed up to $5 \times 10^{-5}$ over the first 500 iterations. We use the pretrained Stable Diffusion v1.5\cite{sd} as the backbone.

To improve model robustness and support both conditional and unconditional colorization, we randomly set the input mask and text to null tokens with a probability of 50\%. All training is performed on 4 NVIDIA A40 GPUs. The first stage is trained for 25,000 iterations, followed by 20,000 iterations in the second stage.

\subsection{Comparison with prior work}

In this section, we qualitatively and quantitatively compare the results generated by our method with those of other state-of-the-art image coloring models. We choose DeOldify \cite{deoldify}, DDColor \cite{DDColor}, CT$^2$ \cite{ct2} and L-CAD \cite{lcad} for unconditional colorization comparison. For fairness, we provided empty text descriptions when testing our model and L-CAD. For all previous methods, we conducted tests using their official codes and weights.

\subsubsection{Quantitative comparison}
\begin{table}[!t]
  \centering
    \renewcommand{\arraystretch}{1.2}
    \caption{Quantitative comparison for unconditional colorization on GPT-color. $\uparrow$($\downarrow$)indicates higher(lower) is better. Best performances are highlighted in \textbf{bold} and second best performances are highlighted in \underline{underline}. Our model performs well on non-reference human perceptual-level metrics.}
    \setlength{\tabcolsep}{3.3mm}{
    \scalebox{0.8}{
    \begin{tabular}{c|cccc|cccc|c}
       \toprule
      \textbf{GPT-Color} &\multicolumn{4}{c|}{Pixel-level metrics} & \multicolumn{4}{c|}{Perceptual-level metrics} & \multirow{2}{*}{Resolution}\\
      Metrics & Colorfulness$\uparrow$ & PSNR$\uparrow$ & SSIM$\uparrow$ & FID$\downarrow$ & NIQE$\downarrow$ & MUSIQ$\uparrow$ & MANIQA$\uparrow$ & TOPIQ\_NR$\uparrow$ & \\
      \midrule
      Deoldify\cite{deoldify} & 25.2940 & \textbf{24.0951} & \textbf{0.9418} & 16.0829 & 3.6085 & \underline{70.1309} & \textbf{0.5018} & 0.5973 & $512\times 512$\\
      DDColor\cite{DDColor} & 35.3415 & \underline{23.7479} & \underline{0.9334} & \textbf{11.0731} & \underline{3.5569} & 69.7063 & \underline{0.4918} & 0.6000 & $512\times 512$\\
      L-CAD\cite{lcad} & 26.8151 & 23.0788 & 0.8837 & 19.8648 & 4.9629 & 56.6726 & 0.4031 & \underline{0.6191} & $256\times 256$\\
      CT$^2$\cite{ct2} & \textbf{40.8147} & 23.0743 & 0.8339 & 12.2452 & 4.6926 & 54.5806 & 0.4347 & 0.5266 & $256\times 256$\\
      \midrule
      Ours & \underline{37.1039} & 23.1224 & 0.8714 & \underline{11.3891} & \textbf{3.5131} & \textbf{70.5013} & 0.4670 & \textbf{0.6234} & $512\times 512$\\
      \bottomrule
    \end{tabular}
    }
    }
    \label{tab:gpt-color}
\end{table}

We benchmark our method against previous methods on GPT-color and report quantitative results in Table \ref{tab:gpt-color}. It is worth noting that the metrics widely used in previous works like PSNR, SSIM and FID \cite{fid} mainly focus on the structural similarity between images. However, since images with a high structural similarity to the original image may not necessarily conform to the natural image distribution and human perception, we believe that using perceptual-level metrics is necessary in the task of image colorization. Thus, we introduce 4 perceptual-level non-reference image quality assessment (NR\_IQA) metrics, NIQE \cite{niqe}, MANIQA \cite{maniqa}, MUSIQ\cite{musiq} and TOPIQ\_NR \cite{topiq} to assess our method. We found that our model did not achieve state-of-the-art performance on metrics that reflect the structural similarity since these metrics do not focus on whether the generated images are colorful or realistic. Moreover, MT-Color's output is of higher resolution than other diffusion-based methods' output, which leads to worse pixel-level metric results. In terms of Colorfulness \cite{colorfulness}, our model performs well, which indicates that our model is able to produce colorful results. On human perceptual-level metrics, our model performs better than other models, indicating that the colorful images generated by our model are more in line with natural distribution patterns and human visual perception.

\subsubsection{Qualitative comparison}

The qualitative comparison results are shown in Figure \ref{qualitative compare}. We observed that DeOldify, as a GAN-based model, suffers from large areas of muted colors and a lack of color variety, resulting in poor visual quality. CT$^2$ and DDcolor, as Transformer-based colorization models, produce more vivid and varied colors but exhibit color bleeding issues. Additionally, these models often apply different colors to the same object, such as the sign in the second row, leading to unrealistic results. Both L-CAD and our model are diffusion-based, whose results exhibit almost no color bleeding, with overall vibrant and natural colors in the images. Our model provides a more diverse color palette, such as the colorful sugar needles on the bread in the first row. Moreover, the resolution of MT-Color's results are fixed to $512 \times 512$, which is clearer than the $256 \times 256$ resolution of L-CAD and CT$^2$.More results and analysis are shown in the appendix.

\subsubsection{Comparison with other diffusion-based methods}

\begin{table}[t]
  \caption{Summary of advantages of MT-Color over existing diffusion-based methods.}
  \centering
  \renewcommand{\arraystretch}{1.2}
  \setlength{\tabcolsep}{6pt} % 调整列间距
  \begin{tabular}{c|cccc}
    \toprule
    Method & Resolution & Pixel-level control& Instance-level control& Strict color binding\\
    \midrule
    Diffusing Colors\cite{diffusingcolor} & 256$\times$256 & $\times$ & $\times$ & $\times$ \\
    Piggybacked\cite{piggy}               & 256$\times$256 & $\times$ & $\times$ & $\times$ \\
    L-CAD\cite{lcad}                                 & 256$\times$256 & \checkmark & \checkmark & $\times$ \\
    \midrule
    Ours                                  & 512$\times$512 & \checkmark & \checkmark & \checkmark \\
    \bottomrule
  \end{tabular}
  \label{novel}
\end{table}

Several recent works leverage the generative power of pre-trained diffusion models for image colorization. However, due to the lack of open-source implementations for many of these methods, we are unable to conduct direct qualitative and quantitative comparisons. Instead, Table~\ref{novel} provides a summary comparison between these approaches and our proposed MT-Color.

A common limitation of diffusion-based colorization models is their inability to preserve fine-grained pixel details, largely due to the inherent stochasticity of the diffusion process. This limitation often restricts the output resolution to $256\times256$. In contrast, MT-Color incorporates a pixel-level mask attention mechanism, enabling effective pixel-level control and significantly boosting the output resolution to $512\times512$.

While our method introduces additional computational overhead which is owing to pixel-space attention, multi-instance sampling, and higher image resolution—these trade-offs are justified by the improvement in precision, instance awareness, and visual fidelity. Moreover, the computational cost can be flexibly reduced by scaling down the resolution when necessary.

\subsection{Ablation study}
\subsubsection{Pixel-level masked attention mechanism}

\begin{figure}
    \centering
    \includegraphics[width=0.6\textwidth]{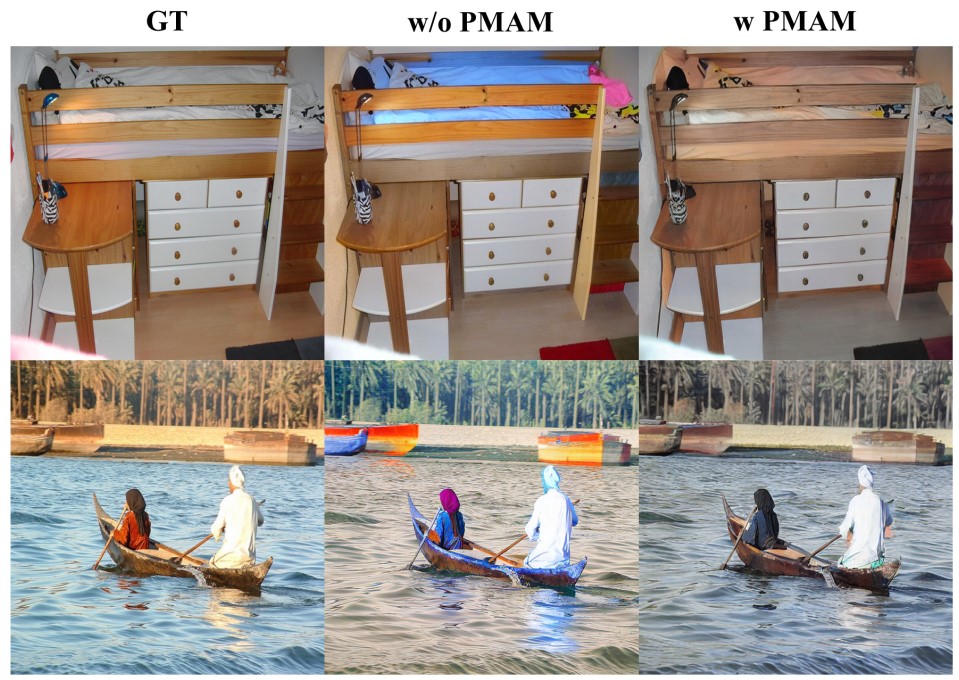}
    \caption{Visual comparison of ablation study for PMAM.}
    \label{ablation-pmam}
\end{figure}

We conduct an ablation study to evaluate the effectiveness of the proposed Pixel-Level Masked Attention Mechanism (PMAM). The quantitative results are summarized in Table~\ref{tab:ablation}, and visual comparisons are presented in Figure~\ref{ablation-pmam}.

\begin{table}[t]
\caption{Quantitative results of ablation study for PMAM.}
\centering
\label{tab:ablation}
\renewcommand{\arraystretch}{1.2}
\begin{tabular}{c|cccc}
\toprule
Method & FID$\downarrow$ & Colorfulness$\uparrow$ & PSNR$\uparrow$ & SSIM$\uparrow$ \\
\midrule
w/o PMAM & 11.88 & 25.69 & 22.71 & 0.8663 \\
w/ PMAM & \textbf{11.39} & \textbf{37.10} & \textbf{23.12} & \textbf{0.8714} \\
\bottomrule
\end{tabular}
\end{table}

By integrating PMAM between ControlNet and the U-Net backbone, MT-Color is able to fully leverage instance-level mask information and enforce precise spatial alignment between conditional features and latent representations. This design \textbf{effectively prevents color spilling beyond object boundaries}, leading to improved visual fidelity. In contrast, removing PMAM results in significant color bleeding and degraded color accuracy, as reflected in both the visual and quantitative results.

\subsubsection{Instance mask and text guidance module}

We evaluate the effectiveness of the instance mask and text guidance module through ablation experiments by comparing the following three model variants:
\begin{itemize}
    \item \textbf{Ours}: The complete model with the full instance mask and text guidance module.
    \item \textbf{Ours w/o mask}: The module is used, but the instance mask is not utilized to construct the attention mask.
    \item \textbf{Ours w/o instance}: The instance mask and text guidance module is entirely removed.
\end{itemize}

\begin{figure}
    \centering
    \includegraphics[width=0.6\textwidth]{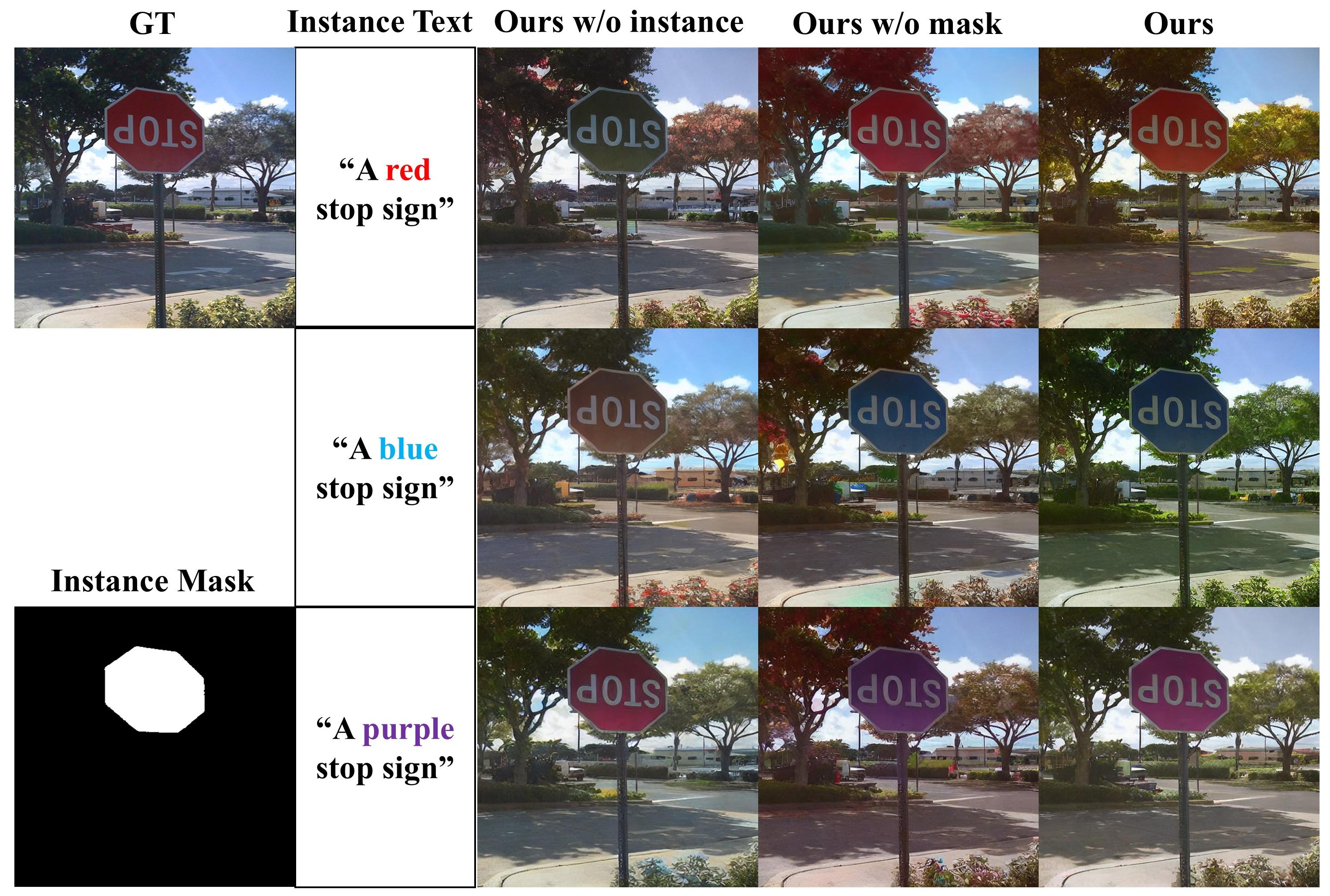}
    \caption{Visual comparison of ablation study on the instance mask and text guidance module.}
    \label{ablation mask and text}
\end{figure}

Qualitative results are shown in Figure~\ref{ablation mask and text}. The instance text format is fixed as \textit{``A \{color\} stop sign''}, where \{color\} represents the target color. We observe that the model without the instance module (\textbf{Ours w/o instance}) fails to correctly apply the specified colors to the stop signs. Although \textbf{Ours w/o mask} can apply the correct color, the absence of attention mask causes color leakage into unrelated regions (e.g., red leaves or purple tints in the background). In contrast, the full model (\textbf{Ours}) accurately binds colors to corresponding objects and confines them strictly within the masked regions, resulting in cleaner and more faithful colorization.

We further conduct quantitative evaluations by computing the CLIP-score \cite{clip} on the GPT-Color test set. Each instance is cropped using its mask, and the CLIP-score is calculated between the cropped region and its corresponding text. As shown in Table~\ref{ablation mask and text table}, the complete model achieves the highest score, indicating stronger alignment between generated colors and textual descriptions.

\begin{table}[t]
\caption{Quantitative results of ablation study on instance mask and text guidance module.}
\centering
\renewcommand{\arraystretch}{1.2}
\begin{tabular}{c|cccc}
\toprule
Method & CLIP-score$\uparrow$ & Colorfulness$\uparrow$ & FID$\downarrow$ & MUSIQ$\uparrow$ \\
\midrule
Ours w/o instance & 0.1944 & 36.28 & \textbf{11.24} & 70.43 \\
Ours w/o mask & 0.2230 & 36.63 & 11.66 & 69.83 \\
Ours & \textbf{0.2273} & \textbf{37.10} & 11.39 & \textbf{70.50} \\
\bottomrule
\end{tabular}
\label{ablation mask and text table}
\end{table}

\subsubsection{Multi-instance sampling strategy}

We also evaluate the effectiveness of the proposed multi-instance sampling strategy using the following variants:
\begin{itemize}
    \item \textbf{Ours}: The full model using multi-instance sampling.
    \item \textbf{Ours w/o crop}: Multi-instance sampling is applied, but the results for each instance are averaged and added to the global result without cropping by instance masks.
    \item \textbf{DDIM}: No multi-instance sampling; instead, standard DDIM is used for denoising.
\end{itemize}

\begin{figure}
    \centering
    \includegraphics[width=0.6\textwidth]{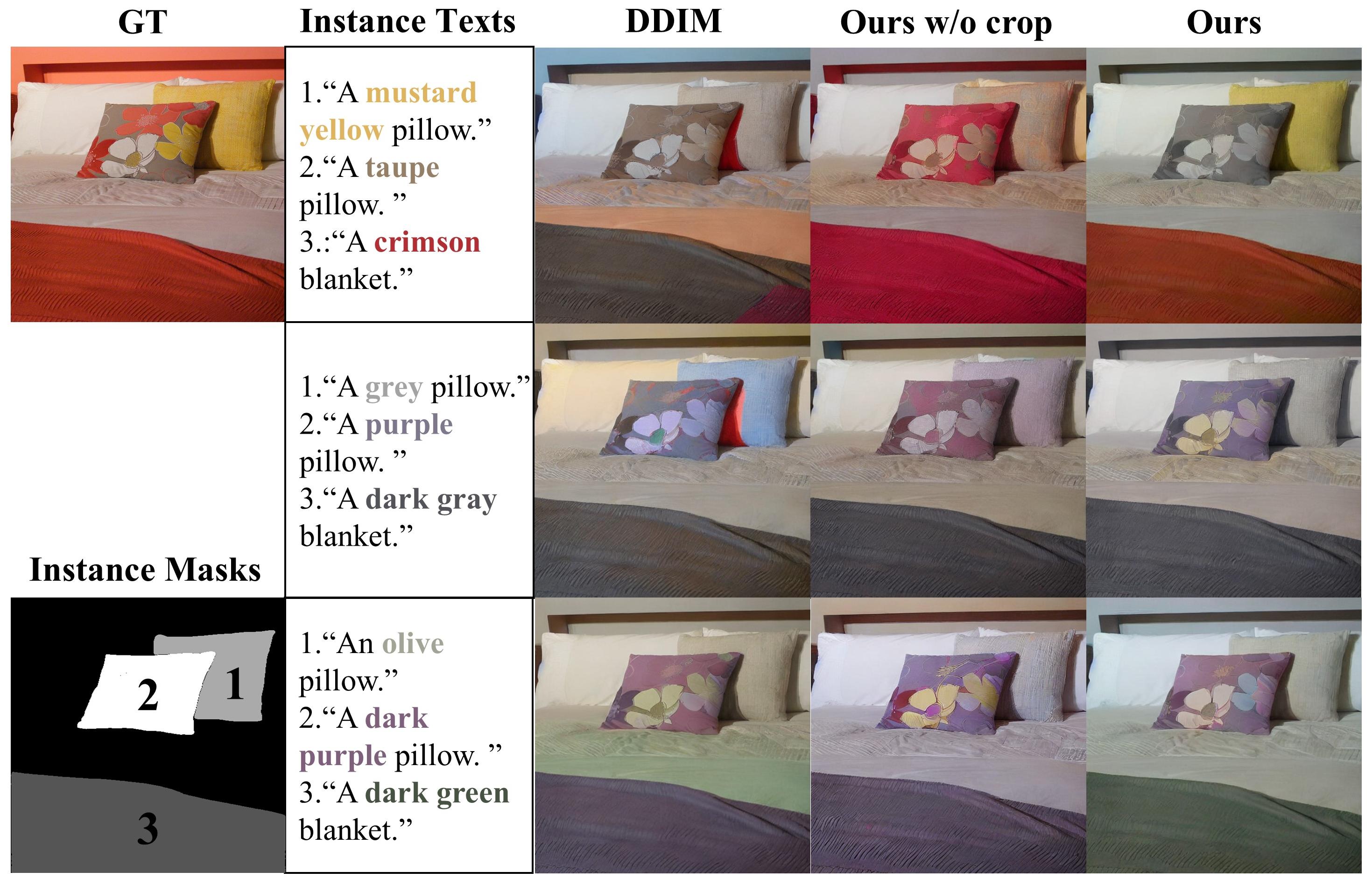}
    \caption{Visual comparison of ablation study on the multi-instance sampling strategy.}
    \label{ablation multi-instance}
\end{figure}

Figure~\ref{ablation multi-instance} shows qualitative comparisons. The baseline DDIM fails to apply the correct colors according to the textual descriptions. The \textbf{Ours w/o crop} variant partially improves results but still suffers from interference between instances. Only the complete method (\textbf{Ours}) correctly assigns the specified colors to corresponding objects while preserving region integrity.

Quantitative results in Table~\ref{ablation multi-instance table} show that the complete multi-instance sampling method achieves the highest CLIP-score and MUSIQ, indicating improved semantic alignment and perceptual quality. These results validate the necessity of separately sampling and fusing instance-level results with region-aware cropping.

\begin{table}[t]
\caption{Quantitative results of ablation study on the multi-instance sampling strategy.}
\centering
\renewcommand{\arraystretch}{1.2}
\begin{tabular}{c|cccc}
\toprule
Method & CLIP-score$\uparrow$ & Colorfulness$\uparrow$ & FID$\downarrow$ & MUSIQ$\uparrow$ \\
\midrule
DDIM & 0.2162 & 36.14 & 11.41 & 68.54 \\
Ours w/o crop & 0.2198 & \textbf{37.25} & 12.23 & 70.16 \\
Ours & \textbf{0.2273} & 37.10 & \textbf{11.39} & \textbf{70.50} \\
\bottomrule
\end{tabular}
\label{ablation multi-instance table}
\end{table}

\subsubsection{Dataset comparison}

\begin{table}
  \centering
  \caption{Comparison between GPT-Color and other datasets.}
  \renewcommand{\arraystretch}{1.2}
  \setlength{\tabcolsep}{5pt}
  \begin{tabular}{c|ccc}
    \toprule
    \multirow{2}{*}{Dataset} & Automatic text & Enhanced color & Instance \\
     & generation & information & text \\
    \midrule
    COCO-Stuff & $\times$ & $\times$ & $\surd$ \\
    Multi-instance & $\surd$ & $\surd$ & $\times$ \\
    GPT-Color & $\surd$ & $\surd$ & $\surd$ \\
    \bottomrule
  \end{tabular}
  \label{GPT-color compare}
\end{table}

In this section, we compare the proposed GPT-Color dataset with other publicly available image colorization datasets that include textual descriptions, to demonstrate its superiority in supporting high-quality colorization models.

Currently, two mainstream COCO-based datasets are used for image colorization: COCO-Stuff and Multi-instance. COCO-Stuff is primarily designed for instance segmentation, where the global text annotations are manually written, and the instance-level annotations are limited to category labels, lacking detailed color information. Multi-instance is tailored for colorization tasks, where global text is generated by BLIP, but it does not provide instance-level textual descriptions. As summarized in Table~\ref{GPT-color compare}, GPT-Color combines the strengths of both datasets—it supports automatic text generation, includes rich color information, and provides fine-grained instance-level descriptions.

\begin{table}[ht]
  \centering
  \caption{Quantitative comparison of textual descriptions among datasets using CLIP-Score.}
  \renewcommand{\arraystretch}{1.3}
  \begin{tabular}{c|c}
    \toprule
    Metric & CLIP-Score $\uparrow$ \\
    \midrule
    COCO-Stuff (Global) & 0.3019 \\
    Multi-instance (Global) & 0.2728 \\
    GPT-Color (Global) & \textbf{0.3059} \\
    \midrule
    COCO-Stuff (Instance) & 0.2115 \\
    GPT-Color (Instance) & \textbf{0.2455} \\
    \bottomrule
  \end{tabular}
  
  \label{gpt-color compare table}
\end{table}

We demonstrate several visual samples for qualitatively comparison in the appendix. To quantitatively assess the quality of textual annotations, we compute the CLIP-Score between global text and images across the three datasets. For instance-level evaluation, we apply instance masks from COCO-Stuff and GPT-Color to extract individual instance regions and then compute the CLIP-Score between the cropped image patches and their corresponding instance descriptions. Results are shown in Table~\ref{gpt-color compare table}.

We observe that Multi-instance yields the lowest global CLIP-Score, likely due to the presence of non-descriptive or irrelevant text such as questions. COCO-Stuff performs better in this regard, but GPT-Color achieves the highest global CLIP-Score, indicating the best overall text-image alignment. For instance-level comparison, GPT-Color also outperforms COCO-Stuff, thanks to its detailed and color-aware instance annotations, which are better recognized by the CLIP text encoder.

\begin{table}[ht]
  \centering
  \caption{Performance comparison of models trained on different datasets. }
  \renewcommand{\arraystretch}{1.3}
  \begin{tabular}{c|cccc}
    \toprule
    Metric & FID $\downarrow$ & Colorfulness $\uparrow$ & PSNR $\uparrow$ & SSIM $\uparrow$ \\
    \midrule
    COCO-Stuff & 13.43 & 34.07 & \textbf{23.41} & \textbf{0.8735} \\
    Multi-instance & 23.63 & 25.79 & 22.20 & 0.8685 \\
    GPT-Color & \textbf{11.40} & \textbf{37.10} & 23.12 & 0.8714 \\
    \bottomrule
  \end{tabular}
  
  \label{gpt-color training compare table}
\end{table}

We further evaluate the training capability of each dataset by training the same model on COCO-Stuff, Multi-instance, and GPT-Color, and testing on the GPT-Color test set. Since Multi-instance does not provide instance-level text, we supply empty instance texts during training for fair. The results are shown in Table~\ref{gpt-color training compare table}.

The model trained on COCO-Stuff performs slightly better in PSNR and SSIM, likely due to its larger scale and broader category diversity. However, the model trained on GPT-Color achieves the best performance in terms of FID and colorfulness, highlighting its superior ability to guide vivid and realistic color generation. These results demonstrate that GPT-Color is better suited for text-guided image colorization tasks.

\section{Conclusion}

In this work, we propose \textbf{MT-Color}, a novel framework designed to address the challenges of color bleeding and inaccurate color binding in pre-trained diffusion-based colorization models. To alleviate color leakage, we introduce a pixel-level masked attention mechanism by integrating Stable Diffusion with ControlNet. To enhance instance-level color fidelity, we propose an instance mask and text guidance module that fuses instance masks and textual descriptions with latent features, alongside a multi-instance sampling strategy to prevent cross-instance information leakage. Furthermore, we construct a new dataset, \textbf{GPT-Color}, using GPT-4 and BLIP-2 to generate fine-grained textual color descriptions and corresponding instance masks. Extensive experiments demonstrate that both the proposed method and dataset significantly improve color accuracy and perceptual quality in text-guided image colorization tasks.

\section{Acknowledgment}
This work is supported by National Natural Science Foundation of China (62271308), STCSM(24ZR1432000, 24511106902, 24511106900, 22DZ2229005), 111 plan (BP0719010),  and State Key Laboratory of UHD Video and Audio Production and Presentation.
 
\appendix

\section{Discussion on hyperparameters}

To explore the effect of the two key hyperparameters $\alpha$ and $\beta$, we conduct a series of experiments and report the results in Table~\ref{tab:hyper}. Here, $\alpha$ controls the portion of individual sampling steps, while $\beta$ adjusts the weight of global noise during sampling.

\begin{table}
\centering
    \caption{Hyperparameters comparison study.}
    \begin{tabular}{cc|ccc}
        \toprule
       $\alpha$ & $\beta$ & CLIP-score$\uparrow$ & PSNR$\uparrow$ & FID$\downarrow$ \\
      \midrule
      0 & 0 & 0.2162 & 22.36 & 11.41 \\
      0.2 & 0 & 0.2240 & 21.07 & 15.87 \\
      0.2 & 0.2 & 0.2273 & \textbf{23.12} & \textbf{11.39} \\
      0.2 & 0.4 & 0.2124 & 22.76 & 13.78 \\
      0.4 & 0.4 & \textbf{0.2305} & 20.81 & 17.76 \\
      \bottomrule
    \end{tabular}
    \vspace{-1em}
    
    \label{tab:hyper}
\end{table}

From the results in Table~\ref{tab:hyper}, we observe that both $\alpha$ and $\beta$ play an important role in balancing semantic alignment, reconstruction fidelity, and realism. We find that higher values of $\alpha$ and lower $\beta$ enhance the binding between instances and texts, but often lead to unstable or inconsistent generation quality. When both hyperparameters are disabled ($\alpha=0, \beta=0$), the model yields moderate performance across all three metrics, serving as a baseline without multi-instance sampling.

Setting both $\alpha$ and $\beta$ to 0.4 results in the highest CLIP-Score (0.2305), but the lowest PSNR and the worst FID (17.76), indicating a significant trade-off: although text-image alignment improves, the generated images become less faithful and perceptually coherent.

When both $\alpha$ and $\beta$ are set to 0.2, the model achieves the best balance: the highest PSNR (23.12) and a competitive FID (11.39), while maintaining a strong CLIP-Score (0.2273).We therefore choose this setting as our default during inference.

\section{Visual dataset comparison}

\begin{figure}
    \centering
    \includegraphics[width=\textwidth]{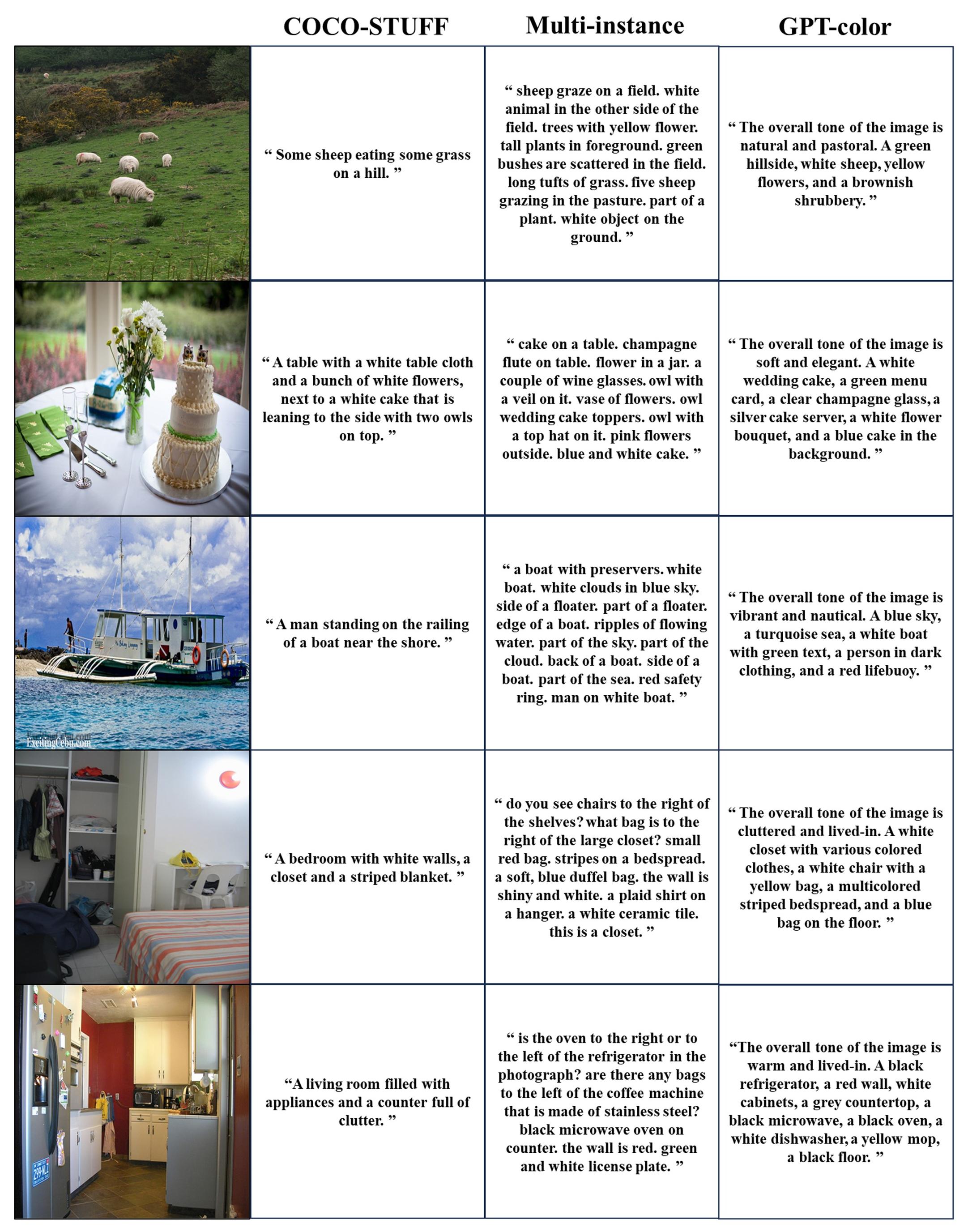}
    \caption{Qualitative comparison of global textual descriptions across GPT-Color, COCO-Stuff, and Multi-Instance datasets.}
    \label{gpt_color_compare_fig}
\end{figure}

To further demonstrate the advantages of GPT-Color in generating textual descriptions, we randomly selected several images from the COCO dataset and compared the corresponding global text descriptions from GPT-Color, COCO-Stuff, and Multi-Instance datasets, as shown in Figure~\ref{gpt_color_compare_fig}. 

We observe that the global text in COCO-Stuff is notably brief, focusing primarily on the general scene and covering only a limited subset of the objects present in the image. Moreover, it lacks detailed color information, which is essential for image colorization tasks. In contrast, the Multi-Instance dataset provides longer descriptions that mention more objects than COCO-Stuff. However, the descriptions often contain irrelevant or non-informative sentences, such as rhetorical questions or repetitive mentions of the same object (e.g., “side of a floater,” “part of a floater,” and “edge of a boat” in the third image). Additionally, despite being tailored for image colorization, Multi-Instance does not consistently provide color details for every mentioned object.

In comparison, GPT-Color begins with a description of the overall color tone of the image, followed by a comprehensive enumeration of objects within the scene, each annotated with specific color information. This structure ensures both completeness and relevance in the text. 

From a qualitative perspective, the textual annotations in GPT-Color are more informative, coherent, and better suited for guiding colorization tasks than those in COCO-Stuff and Multi-Instance.

\section{More visual results of conditional colorization}

In this section, we present several examples from the GPT-Color validation set to further demonstrate the conditional colorization capabilities of our proposed MT-Color model, which leverages global text descriptions, instance segmentation masks, and instance-level textual annotations.

\begin{figure}
\centering
\captionsetup[subfloat]{labelformat=empty} % 移除子图的标签
\subfloat{\includegraphics[scale=0.13]{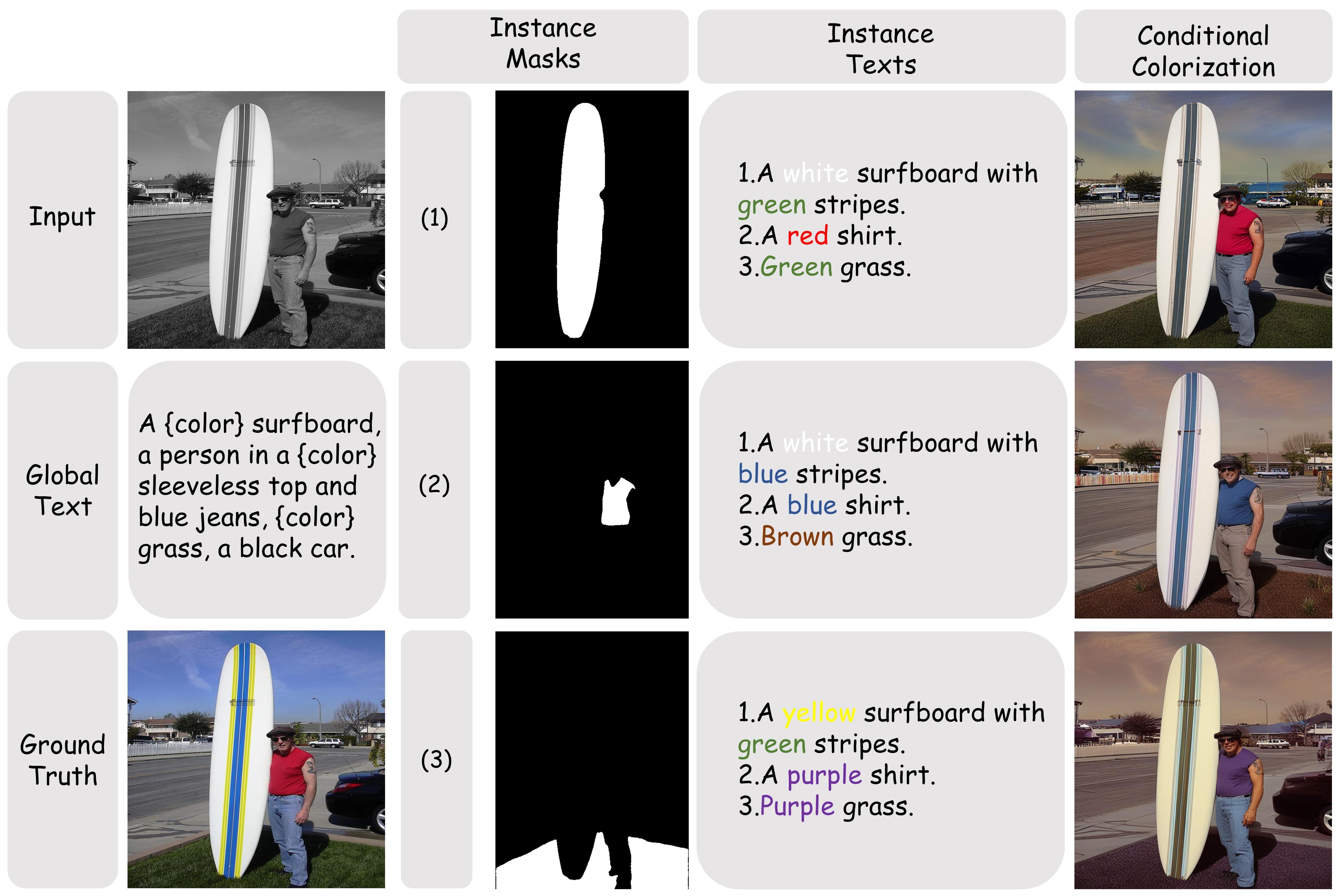}}
\\
\subfloat{\includegraphics[scale=0.158]{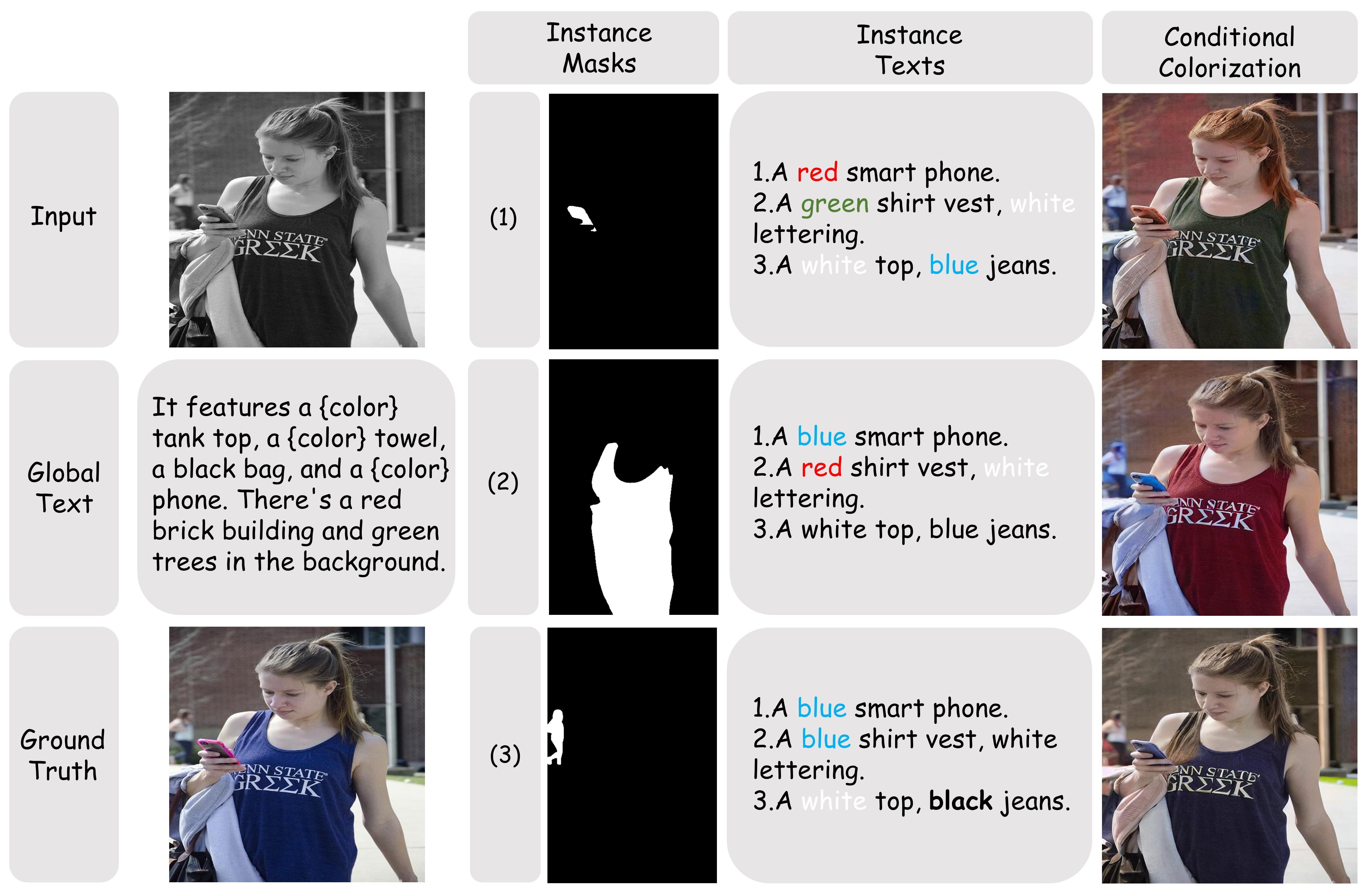}}
\caption{Visual examples of conditional colorization with global texts, instance masks and instance texts on GPT-color.}
\label{cond_show_figure}
\end{figure}

As shown in Figure~\ref{cond_show_figure}, MT-Color is capable of producing precise and diverse instance-aware image colorization guided by user-provided global descriptions, instance masks, and instance texts. The colorization results not only adhere closely to the color constraints specified by the instance-level inputs, but also align well with human visual perception. Furthermore, the output resolution is increased to $512 \times 512$, offering clearer and more visually appealing results.

Benefiting from the multi-instance sampling strategy, the color of each object is influenced not only by the corresponding instance text, but also by the global description. For instance, in the case where the towel held by a girl is not annotated with an instance mask or description, its color is still correctly inferred based on the global text input.

However, due to the inherent stochasticity of diffusion models, MT-Color may occasionally fail to preserve fine pixel-level details from the grayscale input or even generate suboptimal results. Addressing this limitation remains an open direction for future work.

\section{More comparisons of automatic colorization}
\begin{table}
  \centering
   \caption{Quantitative comparison for unconditional colorization on COCO-Stuff dataset. $\uparrow$($\downarrow$) indicates higher(lower) is better. Best performances are highlighted in \textbf{bold} and second best performances are highlighted in \underline{underline}.}
    \renewcommand{\arraystretch}{1.3}
    \begin{tabular}{c|cccc|cccc}
        %COCO-Stuff table
      \toprule
      \textbf{COCO-Stuff} &\multicolumn{4}{c|}{Pixel-level} & \multicolumn{4}{c}{Perceptual-level} \\
      Metrics & Colorfulness$\uparrow$ & PSNR$\uparrow$ & SSIM$\uparrow$ & FID$\downarrow$ & NIQE$\downarrow$ & MUSIQ$\uparrow$ & MANIQA$\uparrow$ & TOPIQ\_NR$\uparrow$\\
      \midrule
      Deoldify & 25.7857 & \underline{23.2442} & 0.8677 & 15.3988 & \underline{3.6135} & \underline{69.9131} & \underline{0.5004} & 0.5961 \\
      DDColor & \underline{35.4759} & 22.9509 & 0.8614 & \textbf{9.7483} & \textbf{3.5588} & 69.5143 & 0.4910 & 0.5981 \\
      L-CAD & 28.8897 & \textbf{24.2191} & \textbf{0.8728} & \underline{11.3773} & 4.9759 & 54.5308 & 0.4021 & \underline{0.6283} \\
      CT$^2$ & \textbf{40.4601} & 23.0503 & 0.8692 & 12.5434 & 4.7118 & 56.3082 & 0.4361 & 0.5278 \\
      \midrule
      Ours & 33.0894 & 23.0729 & \underline{0.8704} & 11.9967 & 4.1312 & \textbf{70.4190} & \textbf{0.5113} & \textbf{0.6358} \\
      \bottomrule
    \end{tabular}
   
    \label{COCO-Stuff}
 \end{table}

\begin{table}
  \centering
      \caption{Quantitative comparison for unconditional colorization on ImageNet dataset. $\uparrow$($\downarrow$) indicates higher(lower) is better. Best performances are highlighted in \textbf{bold} and second best performances are highlighted in \underline{underline}}
    \renewcommand{\arraystretch}{1.3}
    \begin{tabular}{c|cccc|cccc}
        %imagenet table
      \toprule
      \textbf{ImageNet} &\multicolumn{4}{c|}{Pixel-level} & \multicolumn{4}{c}{Perceptual-level} \\
      Metrics & Colorfulness$\uparrow$ & PSNR$\uparrow$ & SSIM$\uparrow$ & FID$\downarrow$ & NIQE$\downarrow$ & MUSIQ$\uparrow$ & MANIQA$\uparrow$ & TOPIQ\_NR$\uparrow$\\
      \midrule
      Deoldify & 25.6872 & \textbf{23.6722} & \textbf{0.9107} & \underline{7.8766} & \underline{4.5813} & \underline{68.7172} & \textbf{0.5314} & 0.6512 \\
      DDColor & \textbf{40.7589} & 22.6318 & \underline{0.8911} & \textbf{5.136} & 4.5918 & 68.7083 & \underline{0.5258} & \underline{0.6520} \\
      L-CAD & 25.2565 & 22.4030 & 0.8690 & 11.0019 & 5.4117 & 58.6934 & 0.4512 & 0.6497 \\
      CT$^2$ & \underline{40.1252} & \underline{23.2567} & 0.8760 & 11.8491 & 5.3477 & 60.4682 & 0.4769 & 0.6006 \\
      \midrule
      Ours & 35.3771 & 22.7644 & 0.8779 & 10.7835 & \textbf{3.9466} & \textbf{69.2562} & 0.5255 & \textbf{0.6556} \\
      \bottomrule
    \end{tabular}

    \label{imagenet table}
 \end{table}

\begin{figure}
    \centering
    \includegraphics[width=0.9\textwidth]{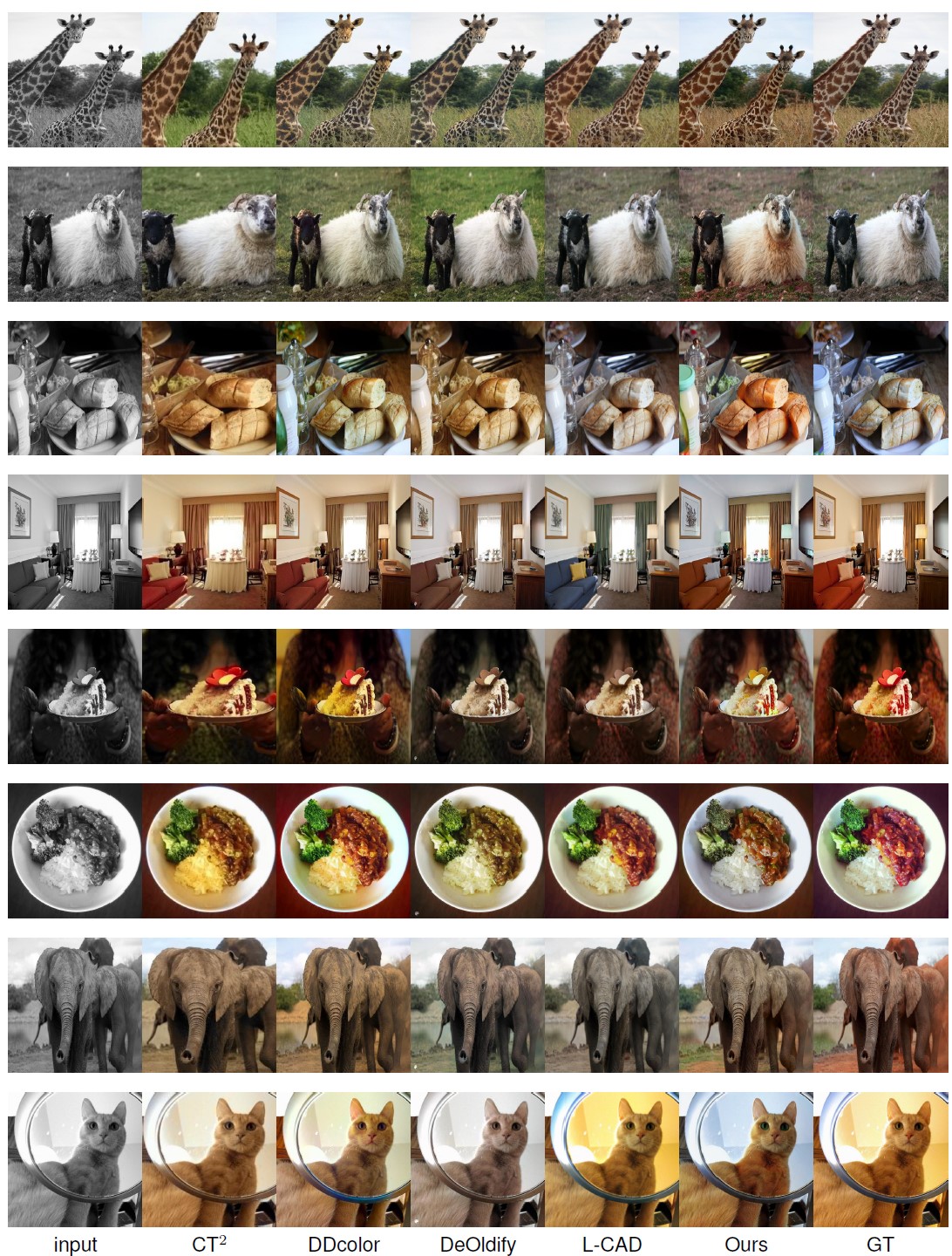} % 插入图片并设置宽度
    \caption{Qualitative comparison results for unconditional colorization. All examples are from COCO-Stuff.}
    \label{coco compare figure}
\end{figure}

\begin{figure}
    \centering
    \includegraphics[width=0.9\textwidth]{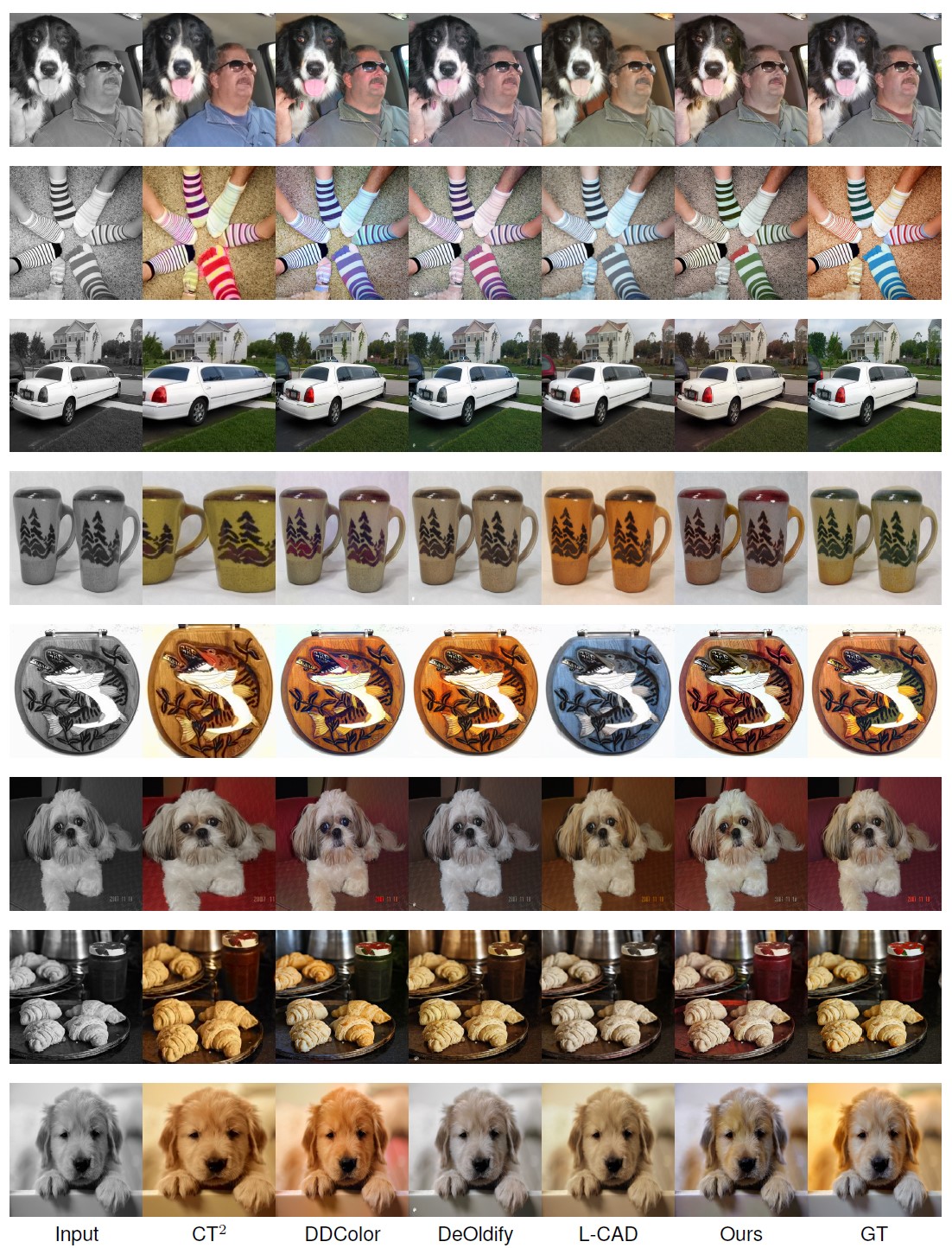} % 插入图片并设置宽度
    \caption{Qualitative comparison results for unconditional colorization. All examples are from ImageNet-5k.}
    \label{imagenet compare figure}
\end{figure}

As shown in Figure \ref{coco compare figure} and \ref{imagenet compare figure}, we provide more unconditional colorization visual results of our model and the comparison with previous methods on COCO-Stuff dataset and ImageNet dataset, respectively. Meanwhile, we test our model and previous methods on these two datasets and report quantitative results in Table \ref{COCO-Stuff} and \ref{imagenet table}.

Since the generated images of CT$^2$ are cropped and resized, in quantitative experiments we crop and resize the ground truth images to match the generated images. As is shown, the results of DeOldify suffer from dull tones and uninspiring colors. Although CT$^2$ and L-CAD could generate colorful and visual appealing images, the resolution of their outputs is limited to $256 \times 256$, which is too low for human visual perception. Since DDColor is based on Transformer architecture and is proposed solely for automatic colorization, it could generate good colorization results while preserving details of the grayscale images and the original resolution. However, DDColor sometimes generate uneven colors due to lack of semantic information. Our proposed MT-Color could not only generate natural and colorful images that better match human visual perception, but also preserve some semantic information learned from the specific-designed dataset and fix the resolution to $512 \times 512$, which is clearer for human perception. Nonetheless, MT-Color sometimes could not preserve pixel details due to the stochasticity of diffusion models.

\section{Discussion on computation costs}

We extend the original ControlNet architecture from latent space to pixel space and adopt a multi-instance sampling strategy during inference to enable precise instance-aware colorization. While these improvements increase computational costs, they significantly enhance performance. Table~\ref{tab:efficiency} presents the computational details of MT-Color and a comparison with L-CAD, a diffusion-based baseline. MT-Color is trained on an NVIDIA A40 GPU, while L-CAD is trained on an NVIDIA RTX 3090 GPU. All inference is conducted using the NVIDIA A40 GPU.

The introduction of a pixel-level attention mechanism and a more sophisticated guidance module significantly increases the number of parameters in MT-Color, nearly doubling those of L-CAD. During inference, MT-Color produces images with a resolution of $512 \times 512$, which contains four times as many pixels as L-CAD’s $256 \times 256$ output. Consequently, the inference time of MT-Color is approximately four times that of L-CAD due to the additional computational cost introduced by pixel-level attention and multi-instance sampling. Nonetheless, this cost is acceptable given the improvements in output resolution and instance-level precision. MT-Color’s memory usage during inference is around 18.0~GB, which remains within the range of high-end consumer GPUs.

\begin{table}
\centering
\caption{Computation details of MT-Color and L-CAD under $\alpha=0.1$.}
\scalebox{1}{
\begin{tabular}{c|c|cc|ccc}
\toprule
\multirow{2}{*}{Method} & \multirow{2}{*}{Parameters} & \multicolumn{2}{c|}{Training} & \multicolumn{3}{c}{Inference} \\
& & Time & Memory & Time per Image & Memory & Resolution\\
\midrule
L-CAD & 1052M & 120h & 14.3GB & 15.2s & 7.3GB & $256\times 256$ \\
MT-Color (Ours) & 1950M & 124h & 40.7GB & 68.6s & 18.0GB & $512\times 512$ \\
\bottomrule
\end{tabular}
}
\label{tab:efficiency}
\end{table}

\printcredits

\bibliographystyle{cas-model2-names}

\bibliography{ref}



\end{document}